\documentclass[sigconf]{acmart}

\usepackage{amsmath,amssymb,amsfonts}
\usepackage{graphicx}
\usepackage{textcomp}
\usepackage{float,array}
\usepackage{mwe,tikz}
\usepackage{amsmath}

\usepackage{amsthm}
\usepackage{float}
\usepackage{enumitem}
\usepackage{hyperref}

\usepackage{rotating}
\usepackage{amssymb}
\usepackage{algpseudocode}
\usepackage{algorithm}
\usepackage{epstopdf}

\usepackage{multirow}
\usepackage{subfig}
\usepackage{dblfloatfix} 

\usepackage[textsize=tiny]{todonotes}
\usepackage{bbm}
\usepackage{xspace}
\usepackage[compact]{titlesec}

\makeatletter
\renewcommand*\env@matrix[1][*\c@MaxMatrixCols c]{%
  \hskip -\arraycolsep
  \let\@ifnextchar\new@ifnextchar
  \array{#1}}
\makeatother

\newcommand\paraspace{\vspace*{0.5ex}}
\newcommand\parab[1]{\paraspace\noindent\textbf{#1}}
\newcommand\parae[1]{\paraspace\textbf{\textit{#1}}}

\newcommand{\etc}{\emph{etc.}\xspace}
\newcommand{\ie}{\emph{i.e.,}\xspace}
\newcommand{\eg}{\emph{e.g.,}\xspace}

\newcommand{\secref}[1]{\S\ref{#1}}
\newcommand{\figref}[1]{Figure~\ref{#1}}

\newcommand{\eqnref}[1]{Equation~\ref{#1}}

\makeatletter
\renewcommand{\verbatim@font}{\ttfamily\bfseries\footnotesize}
\makeatother

\def\BibTeX{{\rm B\kern-.05em{\sc i\kern-.025em b}\kern-.08emT\kern-.1667em\lower.7ex\hbox{E}\kern-.125emX}}

\makeatletter
\def\maxwidth{\ifdim\Gin@nat@width>\linewidth\linewidth\else\Gin@nat@width\fi}
\def\maxheight{\ifdim\Gin@nat@height>\textheight\textheight\else\Gin@nat@height\fi}
\makeatother
\setkeys{Gin}{width=\maxwidth,height=\maxheight,keepaspectratio}
\makeatletter
\g@addto@macro{\UrlBreaks}{\UrlOrds}
\makeatother

\makeatletter
\let\origsection\section
\let\origsubsection\subsection

\renewcommand\section{\@ifstar{\starsection}{\nostarsection}}
\renewcommand\subsection{\@ifstar{\starsubsection}{\nostarsubsection}}

\newcommand\sectionprelude{\vspace{0ex}}
\newcommand\sectionpostlude{\vspace{0ex}}
\newcommand\subsectionprelude{\vspace{0ex}}
\newcommand\subsectionpostlude{\vspace{0ex}}

\newcommand\nostarsection[1]{\sectionprelude\origsection{#1}\sectionpostlude}
\newcommand\starsection[1]{\sectionprelude\origsection*{#1}\sectionpostlude}

\newcommand\nostarsubsection[1]{\subsectionprelude\origsubsection{#1}\subsectionpostlude}
\newcommand\starsubsection[1]{\subsectionprelude\origsubsection*{#1}\subsectionpostlude}

\makeatother

\renewcommand\paraspace{\vspace*{1ex}}
\providecommand\parab[1]{\paraspace\noindent\textbf{#1}}
\providecommand\parae[1]{\paraspace\textbf{\textit{#1}}}

\apptocmd\normalsize{%
\abovedisplayskip=5pt
\abovedisplayshortskip=5pt
\belowdisplayskip=5pt
\belowdisplayshortskip=5pt
}{}{}

\renewcommand\footnotetextcopyrightpermission[1]{} % removes footnote with conference info
\setcopyright{none}
\acmDOI{}
\acmISBN{}
\acmConference[Submitted for review]{}
\acmYear{2019}
\copyrightyear{}
\acmPrice{}
\begin{document}

\title{On Localizing a Camera from a Single Image}

\author{Pradipta Ghosh}
\affiliation{\institution{University of Southern California}}
\email{pradiptg@usc.edu}

\author{Xiaochen Liu}
\affiliation{\institution{University of Southern California}}
\email{liu851@usc.edu}

\author{Hang Qiu}
\affiliation{\institution{University of Southern California}}
\email{hangqiu@usc.edu}

\author{Marcos A. M. Vieira}
\affiliation{\institution{Universidade Federal de Minas Gerais}}
\email{mmvieira@dcc.ufmg.br}

\author{Gaurav S. Sukhatme }
\affiliation{\institution{University of Southern California}}
\email{gaurav@usc.edu}

\author{Ramesh Govindan}
\affiliation{\institution{University of Southern California}}
\email{ramesh@usc.edu}

\renewcommand{\shortauthors}{P. Ghosh et. al.}

\begin{abstract}

Public cameras often have limited metadata describing their attributes. A key missing attribute is the precise location of the camera, using which it is possible to precisely pinpoint the location of events seen in the camera. In this paper, we explore the following question: under what conditions is it possible to estimate the location of a camera from a \textit{single image} taken by the camera? We show that, using a judicious combination of projective geometry, neural networks, and crowd-sourced annotations from human workers, it is possible to position 95\% of the images in our test data set to within 12~m. This performance is two orders of magnitude better than PoseNet, a state-of-the-art neural network that, when trained on a large corpus of images in an area, can estimate the pose of a single image. Finally, we show that the camera's inferred position and intrinsic parameters can help design a number of \textit{virtual sensors}, all of which are reasonably accurate.

\end{abstract}

%%% Local Variables:
%%% mode: latex
%%% TeX-master: "camloc"
%%% End:
\thanks{Research reported in this paper was sponsored in part by the Army Research Laboratory under Cooperative Agreement W911NF-17-2-0196. The views and conclusions contained in this document are those of the authors and should not be interpreted as representing the official policies, either expressed or implied, of the Army Research Laboratory or the U.S. Government. The U.S. Government is authorized to reproduce and distribute reprints for Government purposes notwithstanding any copyright notation here on.}

\maketitle

\section{Introduction}
\label{sec:Introduction}

In the era of millions of smart devices and of emerging smart cities, the camera has become the most ubiquitous sensor deployed on the planet. With its ability to capture images and videos, it can obtain more detail about an environment than almost any other sensor can. As a result, in today's world, there exist many cameras in a typical city like Los Angeles, San Francisco, and New York; a simple web search yields lists of hundreds of these cameras~\cite{abccam, insecam, earthcam}. In a smart city environment, one can envision using these cameras for personalized fine grain navigation~\cite{schoning2009photomap} to overcome GPS errors, for providing fast multi-camera surveillance~\cite{wang2013intelligent} for health care, police personnel, or disaster relief operations.

However, in order to leverage these cameras for such applications, it is important to have accurate meta-data about the camera such as its focal length, its resolution, and its \textit{pose} (location and viewing direction). Unfortunately, even something as simple as the precise location of a camera is not documented in lists of public outdoor cameras~\cite{abccam, insecam, earthcam}. Preliminary analysis reveals that more than 90\% of these sources only have coarse location labels, which vary from city to street level.

%\pradipta{Add more on motivations}

\parab{The Problem.}
In this paper, we address the problem of localizing a camera. The problem of localizing and characterizing a camera from the camera feed is often known in the robotics and vision community (\secref{sec:related_work}) as the \emph{camera-relocalization problem}~\cite{Kendall_2015_ICCV}. In these communities, the problem arises in map construction~\cite{prusak2008pose,sattler2015hyperpoints,Shotton_2013_CVPR}, to estimate the pose of the camera that captures 2D or 3D imagery for constructing the map. Because of their reliance on special sensors, these approaches may not be suitable for surveillance cameras. More recent work has proposed training, using a database of images with known poses, a neural network for camera pose estimation~\cite{Kendall_2015_ICCV}. This approach, too, does not generalize to localizing static surveillance cameras (\secref{sec:experiment}) primarily because there exists no large-scale open-source database of images that captures the diversity of properties and pose in real-world surveillance cameras; for example, the Google street view dataset consists of images taken at the street level, while surveillance cameras are often mounted at different heights.

In this paper, we take a first step towards understanding camera properties in the wild. Our approach is \textit{deliberately} minimalistic. We ask: \textit{Under what conditions is it possible to estimate the location of a camera from a single image taken by the camera?} Of course, just given an image and no additional information, it is difficult to estimate the camera's pose. However, there \textit{is} information (such as a landmark or a road feature such as a street corner) in the image that a neural network or a human might be able to extract. That, together with information from map APIs, might plausibly be able to localize the camera. In this paper, we explore \textit{\textbf{the smallest set of additional information needed to localize the camera to within meter-level accuracy, given just a single image}}. We have left to future work to understand how to leverage \textit{multiple} images to improve camera localization; our results provide a upper bound that future work can attempt to improve.

%% (2) Such system mainly
%% performs well if both training and testing images are taken from
%% cameras with similar properties. This leads us to focus on the
%% following revised questions: \emph{under what conditions is it
%%   possible to estimate the location of a camera from a single image
%%   taken by the camera without requiring a 3D map of the environments
%%   and resource-heavy time consuming training? Can we develop a method
%%   that is generalizable to any heterogeneous set of camera?}

% \hang{the focal length is a bit abrupt, out of context here.}

\parab{Approach.}
To localize a camera from a single image, we adapt techniques from projective geometry (\secref{sec:background}), and methodically develop a set of algorithms that combine these with a judicious combination of neural networks and crowd-tasking. We find that in order to estimate the camera location from a single image, we need to first solve for a total of eighteen unknowns corresponding to: five \textit{intrinsic} properties of the camera including the focal length of the camera, the pixel dimensions of the image, and the location of the image's principal point \ie image center; twelve unknowns corresponding to \textit{extrinsic} properties such as position and the orientation of the camera (\secref{sec:rel_loc}); and one special unknown corresponding to the homogeneous camera projection equation (\secref{sec:rel_loc}). Solving these unknowns can provide us the relative location of the camera with respect to a reference point visible in the image. To get the mapping from the relative location to the absolute location, we need to know the absolute (\ie GPS) location of the reference point.

%% \added{For illustration, we present two applications of using a single street facing camera for gathering information about speed of the vehicles and blind people navigation that would be impossible without accurate meta-data. Future smart city disaster relief operations will likely to involve multiple cameras spanning across the region. To this, one would require to select an optimal subset of cameras to fulfil the objective that would also require the precise location and properties of a camera.}

%% Now the
%% questions are: \emph{how can we estimate the values of these seventeen
%%   unknowns from the image?} \emph{how can we get the pixel location to
%%   GPS location mapping for a reference point from the image?}
%\pradipta{ we should talk about the multiple images a bit?}

\parab{Contributions.} Our paper makes three important contributions.

\parae{Camera Relative Position.} To estimate the five unknown camera intrinsic parameters, we develop novel image annotation methods to permit the use of vanishing point detection techniques from projective geometry~\cite{rother2002new}. The vanishing point (VP) of an image refers to the point or direction in the image to which a set of parallel lines in the real world converges (\figref{fig:vp_illustr}). A set of three orthogonal vanishing points of the image can help estimate the camera intrinsic properties as well as the unknowns related to the orientation of the camera (\secref{sec:rel_loc}). However, the task of finding three orthogonal vanishing is not a trivial one~\cite{simon2016simple}. Automated approaches for this often either require some property of the camera to be known~\cite{6385802} or output a wrong set of orthogonal VPs. We propose an approach that uses annotations from human workers (using a programmable crowd-tasking platform\cite{Satyam}), which leverages human perception and the wisdom of the crowds to achieve better results than possible with automated methods. 

Given the vanishing points, we propose a novel algorithm to estimate the camera's extrinsic properties (position and orientation). This technique needs an annotation for the dimensions of an object visible in the image, such as a car (\secref{sec:rel_loc}). For this, we explore a range of approaches including using a pre-trained model of neural network for object detection, standard object dimensions, or crowd-sourced annotations of the object. We further show that, having estimated all eighteen unknowns, we can estimate the relative location of the camera with respect to any pixel of the image that corresponds to a point on the earth surface. %\pradipta{@Ramesh:Add more on this work not being just engineering work?}

% (\secref{sec:rel_loc}).

\parae{Camera Absolute Position.} To get the absolute position (\ie the GPS coordinate) of the camera, ideally, we need the GPS coordinate of one reference pixel point. However, it turns out (\secref{sec:abs_loc}) that due to ambiguity in the direction between the image and the real world, this is insufficient: we need the GPS coordinates of at least two reference pixels (\secref{sec:abs_loc}). Using the fact that coarse location tags are often available for public cameras, we explore two particular scenarios to obtain positions of reference pixels: (1) when the nearest road intersection to a camera is known, and (2) when a landmark building is visible in the image and can be uniquely identified. For these cases, we develop a complete characterization the solution quality; in each case, we are able to narrow down the absolute location to one of a small number of candidate positions.

\parae{Applications.} Finally, we show that the camera pose and intrinsic parameter estimates can help develop a variety of novel \textit{virtual sensors} using the camera (\secref{s:applications}, \secref{s:applications-1}). A \textit{virtual scale} can measure lengths in the physical world from the camera (\eg the length of a road divider), a \textit{virtual clinometer} can measure building height, and a \textit{virtual radar} can track vehicle speed. Finally, a camera's absolute location can help develop a \textit{virtual guide} for visually-challenged pedestrians.

\parae{Evaluation.}
To evaluate our algorithms, we have developed a software tool called CamLoc (\secref{sec:experiment}) which implements these algorithms, and have evaluated it on a representative dataset of 214 images. Our evaluation (\secref{sec:experiment}) shows that CamLoc can estimate the relative position of the camera with less than 10 meters error in position and less than 3 meters error in height in $\approx 95\%$ of the test cases. Moreover, CamLoc is 1-2 orders of magnitude better than PoseNet~\cite{Kendall_2015_ICCV}, a deep neural network for pose estimation. We further show that if the pixel to GPS mapping of two points are known, CamLoc can output the absolute location of the image with less than 12 meters error in $95\%$ of the test cases. A careful analysis of the images with bad performance of CamLoc shows that the errors are due to bad annotations for VPs as a result of non-obvious three perpendicular directions. We further demonstrate that even in the presence of ambiguity in the available information, CamLoc's output solution almost always contains a candidate location for the camera that is within 10 meters of the actual location of the camera. Finally, we demonstrate that CamLoc's virtual sensors are, in general, accurate to with 10-15\% with high percentile, suggesting that they can help with \textit{triage} (\eg determine where to deploy more reliable speed sensors).

% \marcos{We should mention self-calibrating and heterogeneous cameras somewhere.}

%%% Local Variables:
%%% mode: latex
%%% TeX-master: "camloc"
%%% End:

\section{Background and Approach}
\label{sec:background}

In this section, we discuss background material on modeling cameras
using projective geometry, then describe our overall approach.

\begin{figure*}[!ht]
    \centering
    \subfloat[]{ \label{fig:projection_ill} \includegraphics[width=0.33\linewidth]{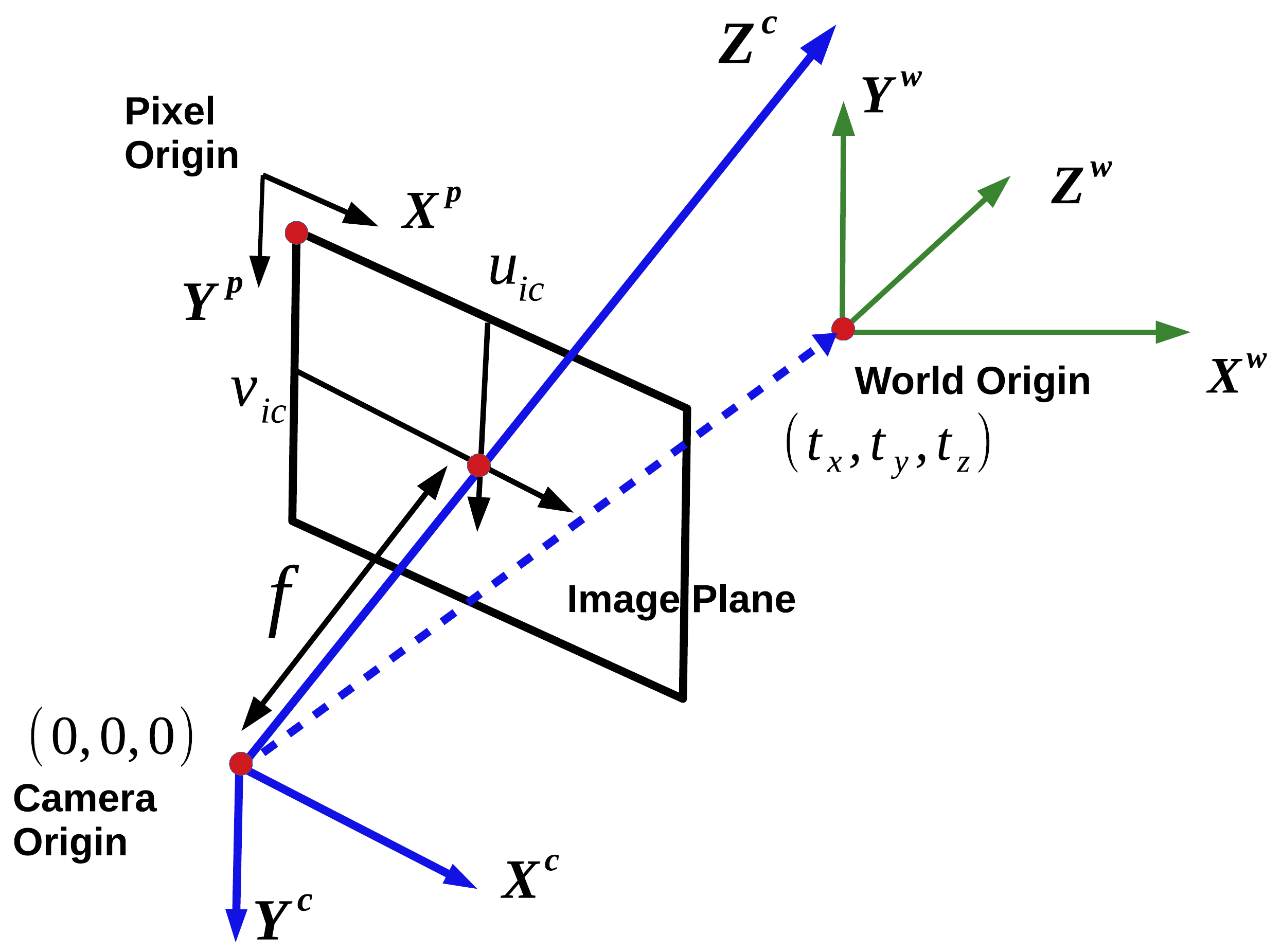}}
    \subfloat[]{\label{fig:vp_illustr} \includegraphics[width=0.3\linewidth]{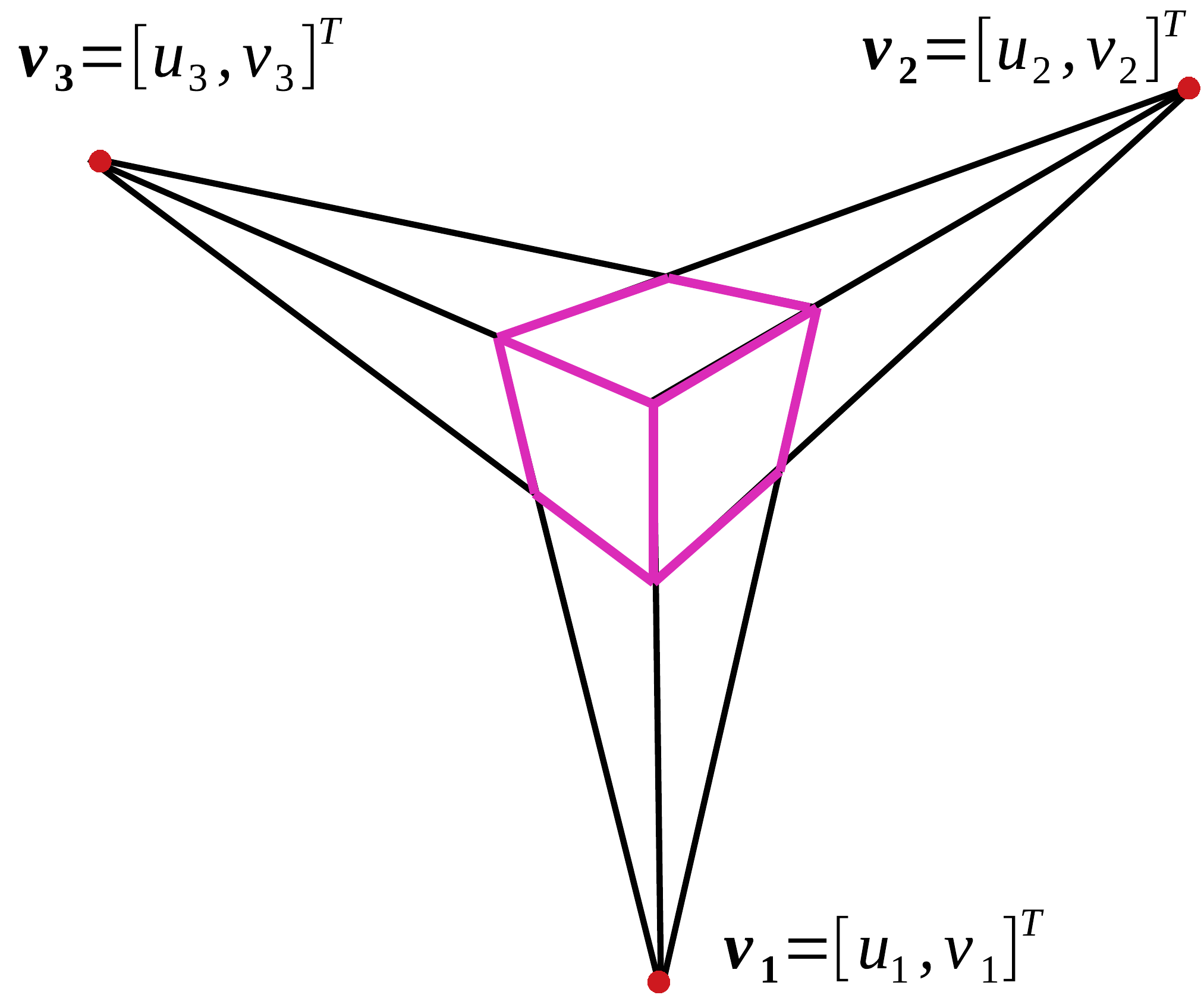}}\qquad
    \subfloat[]{\label{fig:caraxis}\includegraphics[width=0.23\linewidth]{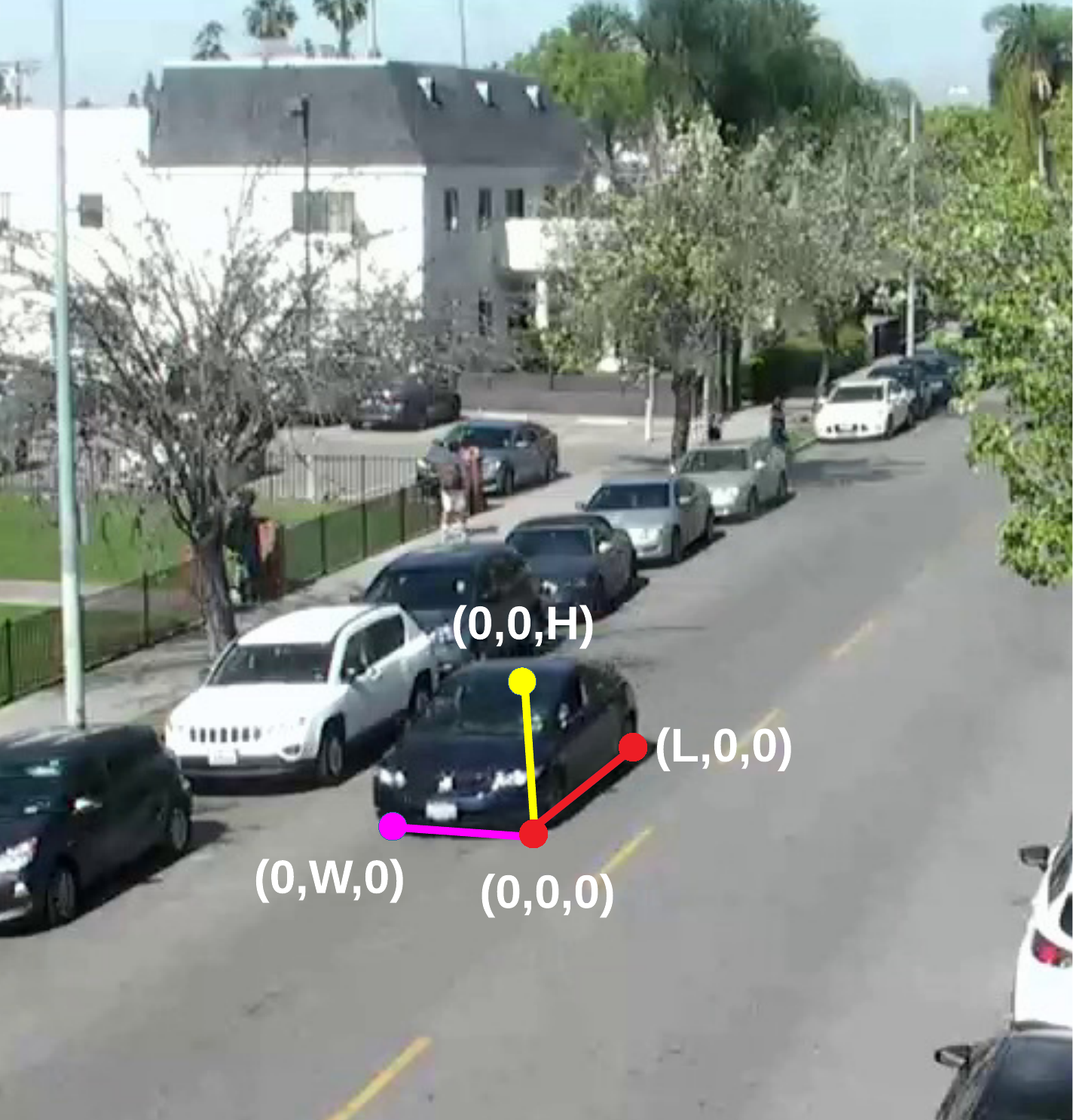}}
    \caption{(a) Illustration of different coordinate systems. (b) Three orthogonal vanishing points for a right-angled cube. (c) Illustration of the world coordinate with the world origin at the nearest bottom corner of the car. L, W, H refer to the car's length, width, and height, respectively.}
    % \label{fig:vp_illustr}
\end{figure*}

\parab{Coordinate Systems.} A point seen by a camera can be
represented in three different coordinate systems: \textit{world},
\textit{camera}, and \textit{pixel}.
%% I commented this out because it isn't used in this section. Let
%% $O^w$ denote the origin of the global coordinate system.
The camera coordinate system has its origin at the camera's focal point with $z$-axis towards the camera's optical axis and the $x$-$y$ direction along the camera image plane (\figref{fig:projection_ill}). The image plane is at a distance $f$ from the camera origin along the $z$ direction of the camera coordinate system. The pixel coordinate system lies on the image plane with origin at the top left corner of any image as illustrated in \figref{fig:projection_ill}. Image pixel locations are always represented in pixel coordinates. The point where the optical axis intersects the image plane is the image center whose pixel coordinates are $\begin{bmatrix} u_{ic} & v_{ic}\end{bmatrix}^T$ and camera coordinates are $\begin{bmatrix} 0 & 0& f\end{bmatrix}^T$.

\parab{Notation.} We denote the concatenation of two matrices $\mathbf{A}$ and $\mathbf{B}$ with same number of rows by $\begin{bmatrix}[c|c]\mathbf{A} & \mathbf{B}\end{bmatrix}$. We also denote the concatenation of two column vectors $\mathbf{a}$ and $\mathbf{b}$ by $\begin{bmatrix}[c||c]\mathbf{a} & \mathbf{b}\end{bmatrix}$.
 
% \begin{figure}
%     \centering
%     \includegraphics[width=0.7\linewidth]{images/projection_2.eps}
%     \caption{Illustration of Different Coordinate Systems}
%     \label{fig:projection_ill}
% \end{figure}

%% Let $[t_x, t_y, t_z]^T$ be the
%% location of the world origin in the camera coordinate system.

\parab{Modeling Cameras using Projective Geometry.}
Two matrices suffice to model any monocular camera: an \emph{extrinsic matrix} and an \textit{intrinsic matrix.} The extrinsic matrix $\mathbf{M} = \begin{bmatrix}[c|c]\mathbf{R} & \mathbf{T}\end{bmatrix}$ contains information about the camera's position and orientation (roll, pitch, and yaw) with respect to the world coordinate system. $\mathbf{R}$ and $\mathbf{T}$ are the rotational and translation matrix of the camera, respectively. $\mathbf{R}$ is a special orthogonal matrix such that $\mathbf{R}\mathbf{R}^T = I_3$ where $I_3$ is a $3\times 3$ identity matrix. This implies $\mathbf{R}^{-1} = \mathbf{R}^T$. Given a point $\mathbf{x}^w$ in the world coordinate system\footnote{To be precise, $\mathbf{x}^w = \begin{bmatrix} x^w & y^w & z^w\end{bmatrix}^T$}, we can determine its camera coordinates using:
\begin{equation}
  \label{eq:camprojection}
  \mathbf{x}^c = \mathbf{M} 
  \begin{bmatrix}[c||c] \mathbf{x}^w &  1\end{bmatrix}
\end{equation}
We call this the \textit{camera projection} equation.

%% \begin{equation}
%%     \mathbf{M} = \begin{bmatrix}[c|c]\mathbf{R} & \mathbf{T}\end{bmatrix} = 
%%     \begin{bmatrix}[ccc|c]\mathbf{r_1} & \mathbf{r_2}& \mathbf{r_2} & \mathbf{T}\end{bmatrix} =
%%     \begin{bmatrix}
%%         r_{11} &r_{12}   &r_{13} & t_x \\
%%         r_{21} &r_{22}  &r_{23} & t_y \\
%%         r_{31} &r_{32}   &r_{33} & t_z \\
%%     \end{bmatrix}
%% \end{equation}
%% where
%% $r_{11}, r_{12}, r_{13}, r_{21}, r_{22}, r_{23}, r_{31}, r_{32},
%% r_{33}$ represent the camera pose in terms of roll, pitch, and yaw
%% which are basically the directions of the world-axes, $X^w$, $Y^w$,
%% $Z^w$, with respect to the Camera Axes $X^c$, $Y^c$, $Z^c$. On the
%% other hand, $t_x$, $t_y$, $t_z$ represent the position of the world
%% origin in camera coordinates. In brief, $\mathbf{M}$ applies necessary
%% rotation and translation to a point in global coordinate
%% $\mathbf{X}^w$ to get the respective coordinate in $\mathcal{R}^c$,
%% $\mathbf{X}^c$.

The intrinsic matrix $\mathbf{K}$ projects a point with camera coordinates $\mathbf{x}^c$ onto the image plane. It requires three parameters:
(a) focal length $f$ of the camera; 
(b) pixel width $dp_x$ and height $dp_y$,  the physical dimension of each pixel; and (c) the image center (described above). Then, to obtain the pixel coordinates $\mathbf{x}^p$ of a point with world coordinate $\mathbf{x}^w$, we use:
\begin{equation}
  \label{eq:pixprojection}
  \lambda  \begin{bmatrix}[c||c] \mathbf{x}^p &  1\end{bmatrix} = \mathbf{K}\mathbf{x}^c = \mathbf{K} \mathbf{M}
    \begin{bmatrix}[c||c] \mathbf{x}^w  &  1\end{bmatrix}
\end{equation}
We call this the \textit{pixel projection} equation. $\lambda$ is a constant used for homogeneous representation of the originally non-linear projection equation~\cite{proj-tutorial}.

\parab{Vanishing Points.}
Our paper estimates the extrinsic and intrinsic matrices using the idea of \emph{vanishing points}. Two parallel lines in the world coordinate system, when projected onto the image plane, appear to converge at a single finite \emph{vanishing point}. However, if those lines are also parallel in the image plane, the vanishing point for those lines is at infinity. Two vanishing points are orthogonal if the parallel lines in the world coordinate system of one are perpendicular to the parallel lines of another. An image can have at most three mutually orthogonal vanishing points (\figref{fig:vp_illustr})~\cite{vanish-tutorial}.

% \hang{\figref{fig:vp_illustr} is kinda confusing. Maybe painting the cude into a different color would help a lot, so the reader immediately know how are the vanishing points drawn.}

%% Vanishing Point is a key concept in projection geometry. Projections
%% of a set of parallel lines in real world i.e., in world coordinate
%% converges to either a finite point in the image plane or at an
%% infinite point as illustrated in Fig.~\ref{fig:vp_illustr}. This point
%% is referred to as the \emph{Vanishing Point for that set of parallel
%%   lines.} The respective direction is referred to as the
%% \emph{``Vanishing Direction''}. The infinite vanishing point exists
%% only if the parallel lines are also parallel to the image plane.

% \begin{figure}[!ht]
%     \centering
%     \includegraphics[width=0.65\linewidth]{images/vanish_point_example.eps}
%     \caption{Three Orthogonal Vanishing Points for a Right-Angled Cube}
%     \label{fig:vp_illustr}
% \end{figure}

We use three properties of vanishing points to estimate the intrinsic and extrinsic matrices of a camera. (1) All lines in the same direction in the real world share the same vanishing point. (2) Given 3 finite orthogonal vanishing points (VP), the image center is the orthocenter of the triangle formed by the vanishing points. (3) Given 2 finite orthogonal vanishing points and one infinite vanishing point, one of the camera axis is parallel to one of the world axis.

%\section{Problem Formulation}

%\ramesh{Move this text to another file to reduce clutter in the top-level file?}

\parab{Approach.}
In this paper, we explore techniques to determine both the \textit{relative} (\secref{sec:rel_loc}) and the \textit{absolute} positions (\secref{sec:abs_loc}) of a street-facing web-camera from which we have a single image. Clearly, using nothing else but the image, it is impossible to determine these positions, so we ask: What is the minimal set of additional contextual information necessary in order to accurately estimate these positions?

Our approach explores two types of contextual information: (a) image \textit{annotations} obtained either through \textit{\textbf{crowd-sourcing}} or using a trained \textit{\textbf{neural network}}, (b) coarse absolute positions such as the nearest landmark or the nearest intersection.

Using these, we first estimate the vanishing points and use these to determine the extrinsic and intrinsic parameters of the camera (\secref{sec:rel_loc}). Then, from the contextual information, we use absolute locations of landmarks or street corners to fix the absolute location of the camera (\secref{sec:abs_loc}). The following sections describe these contributions in greater detail.

%% In this paper, we are interested in the problem of
%% localizing public street-facing web-cameras and estimate their
%% intrinsic and extrinsic matrix parameters based on the available
%% information as well as the accessible camera feeds. Let us first
%% assume that the unknown GPS coordinate of the camera is
%% $(lat_{c}, long_{c})$ and the camera is mounted at a height of $h$.
%% Let us further assume that we know the actual GPS location of two
%% image pixels corresponding to $\mathbf{X}_{rf1}^p $,
%% $\mathbf{X}_{rf2}^p $ as $(lat_{rf1}, long_{rf1})$ and
%% $(lat_{rf2}, long_{rf2})$ respectively. We will refer such points as
%% the \emph{``Reference Points''}. We further assume that the camera is
%% a street facing camera with active traffic. Our overall goal is to
%% estimate the GPS coordinate of the camera as well as the intrinsic and
%% extrinsic parameters of the camera, by leveraging the accessible
%% camera feed and other publicly available datasets and APIs.

%% This objective translates to three different sequential
%% sub-objectives: (1) Estimate the relative position of the the camera
%% and the reference points in terms of meters i.e,
%% $\mathbf{X}^w_{rel} = (\delta x^w, \delta y^w, \delta z^w)$ in the
%% world coordinate system, $\mathcal{R}^w$. .

%% (2) Convert the relative position of the the camera and the reference
%% points into GPS coordinate different, $(\delta_{lat}, \delta_{long})$.

%% (3) Get two of more reference points in order to estimate the real
%% locations.

%%% Local Variables:
%%% mode: latex
%%% TeX-master: "camloc"
%%% End:
%%\input{problem.tex}
\section{Relative Localization}
\label{sec:rel_loc}

This section describes how we can estimate the relative position of a camera from a single image. First, we explain the theory underlying position estimation, some of which we have developed. This description identifies a \textit{sufficient} set of annotations required to estimate the position. We then describe how we obtain these annotations.

\subsection{Estimating Relative Position}
\label{sec:estim-relat-posit}

\parab{Overview.}
We estimate the relative position with respect to world coordinates. So, the first annotation we need is to fix a point in the image as the world origin. Denote this by $\mathbf{x}^w_O$.

Then, to find the position of the world origin in the camera coordinate system, we can use the camera projection equation (\eqnref{eq:camprojection}): $\mathbf{x}^c_O = \mathbf{M} \begin{bmatrix}[c||c] \mathbf{x}^w_O & 1\end{bmatrix}$. Now, the position of the camera in world coordinates is $-\mathbf{R}^T \mathbf{x}^c_O$ where $\mathbf{R}$ is the camera rotational matrix (\secref{sec:background}). But this requires the extrinsic matrix $\mathbf{M}$.

In practice, we can obtain the \emph{pixel coordinates} $\mathbf{x}^p_O$ of the world origin, for example, by a human selecting a specific pixel as the world origin. The camera coordinates of the world origin are, from \eqnref{eq:pixprojection}: $\lambda \begin{bmatrix}[c||c] \mathbf{x}^p_O & 1\end{bmatrix} = \mathbf{K} \mathbf{x}^c_O$. In this case, without knowing the intrinsic matrix $\mathbf{K}$ and $\lambda$, we can not estimate $\mathbf{x}^c_O$.

To estimate the relative position, we need to solve for the extrinsic matrix $\mathbf{M}$ and the intrinsic matrix $\mathbf{K}$ by leveraging the pixel projection equation (\eqnref{eq:pixprojection}):
$\lambda \begin{bmatrix}[c||c] \mathbf{x}^p_O & 1\end{bmatrix} = \mathbf{K}\mathbf{M}\begin{bmatrix}[c||c]  \mathbf{x}^w_O &  1\end{bmatrix}$(\secref{sec:background}). 

Thus, the key challenge in our work is to obtain the extrinsic matrix $\mathbf{M}$ and the intrinsic matrix $\mathbf{K}$. Together, these matrices have the following unknowns:
(a) Three unknowns each for roll, pitch, yaw, and translation;
(b) The camera's focal length ($f$), the 2 pixel dimensions, and the pixel
  coordinates of the image center; and
(c) The unknown constant, $\lambda$.
%   However, in the intrinsic matrix $\mathbf{K}$, the actual quantities of interest are $f_x = f \cdot dp_x$ and $f_y = f \cdot dp_y$. So, it reduces the unknowns to $f_x$, $f_y$ and the pixel
%   coordinates of the image center.
However, $\lambda$ is a function of the seventeen other unknowns mentioned above as well as the global coordinate of the point.

In the rest of this section, we show that, \textbf{\textit{with annotations from which we can extract the following pieces of information, we can estimate the extrinsic and intrinsic matrices, and therefore the camera's relative position}}. The required pieces of information are:
(a) The world origin as well as three orthogonal axes defining the
  world coordinate system;
(b) Three orthogonal vanishing points that correspond to the three axes
direction of the world coordinate system.
(c) The physical length of some object in the environment.

We first describe the mathematics underlying the estimation of these parameters, then discuss how we obtain these annotations.

%% \ramesh{Worth thinking about removing the intrinsic matrix estimation, if it is not our contribution?}

\parab{Estimating the intrinsic matrix $\mathbf{K}$.}
Prior work~\cite{willson1994center} has described how to estimate the intrinsic parameters using a set of three orthogonal vanishing points, which we use directly. \textit{\textbf{Our contribution is the design of annotations to obtain a correct set of orthogonal vanishing points}} (\secref{sec:annotations}). We omit a description of this estimation for brevity.

\parab{Estimating the extrinsic matrix $\mathbf{M}$.} The extrinsic matrix has two parts: the rotational matrix which we estimate using the vanishing points, and the translation matrix which we estimate using the world axes and the object dimension.

\parae{Rotational matrix.} The rotational matrix consists of three vector components $\mathbf{R} = [\mathbf{r_1}\ \mathbf{r_2}\ \mathbf{r_3}] $. Let $\mathbf{v_i}$ denote the $i$-th vanishing point $ i \in \{1, 2, 3\}$. Then, we can use the estimated intrinsic matrix:
\begin{equation}
    \lambda_i \mathbf{K}^{-1}\begin{bmatrix}[c||c] \mathbf{v}_i & 1 \end{bmatrix}  = \mathbf{r}_i 
\end{equation}
This is because each direction of rotation (yaw, pitch and roll) corresponds to one vanishing direction.

To estimate the value of $\lambda_i$, we use the fact that $||\mathbf{r}_i||^2 = 1$~\cite{orghidan2012camera}. Using all three equations related to three vanishing points, we can estimate
% the $\mathbf{r_i}$s, 
all three components of $\mathbf{R}$, modulo one important detail. A rotational matrix must satisfy two properties. In the singular value decomposition (SVD) of the rotational matrix $\mathbf{R} = U\cdot S\cdot V^H$, $S$ should be an identity matrix. Moreover, the determinant of $\mathbf{R}$ should be 1. Since the estimated rotational matrix $\mathbf{R}$ might not satisfy these properties, we sanitize it using a standard SVD decomposition method from image processing~\cite{prince2012computer}.

% \ramesh{Add citation to this sanitizing method.}
% \pradipta{I learned it from an lecture. Was not sure what ref to put. So I just put reference to a computer vision book.}
%% . In order to sanitize it, first, we
%% perform the SVD decomposition of $\mathbf{R}$ and then replace $S$
%% with an $3\times 3$ Identity matrix $I_3$. Then we reconstruct the
%% rotational matrix $\mathbf{R}' = U\cdot I_3\cdot V^H$. Secondly. we
%% check the determinant of $\mathbf{R}'$. If the value is negative, we
%% flip the signs of all the element. This is a standard method of
%% sanitizing rotational matrix in the image processing literature.

%\subsubsection{Translational Matrix Parameter Estimation}

\parae{Translation Matrix.} We are now left with estimating the location of the world coordinate origin in the camera coordinate system, which is same as the camera's translation matrix
$\mathbf{T} = \begin{bmatrix} t_x & t_y & t_z \end{bmatrix}^T$. For this, we need to fix the world coordinate origin $\mathbf{x}^w_O = \begin{bmatrix}0 & 0& 0\end{bmatrix}^T$. Let the corresponding pixel coordinate be $\mathbf{x}^p_O = \begin{bmatrix}u_0 & v_0\end{bmatrix}^T$.

To find $\mathbf{T}$, we need world coordinate of one other point. As we describe in \ref{sec:annotations}, we obtain this from an image annotation: we leverage the fact that there exists, in surveillance cameras, objects with standard (or computable) physical dimensions, like a car (\figref{fig:caraxis}), human, or a common landmark (\eg fire-hydrant). Let this second point be $\mathbf{x}^w_1 = \begin{bmatrix}L & 0& 0\end{bmatrix}^T$, whose corresponding pixel coordinate is $\mathbf{x}^p_1 = \begin{bmatrix}u_1 & v_1\end{bmatrix}^T$.

We can then write the following equations using the intrinsic and extrinsic matrices:
\begin{equation}
\begin{split}
    &\lambda_0 \mathbf{K}^{-1}\begin{bmatrix}[c||c]\mathbf{x}^p_O & 1 \end{bmatrix}   = \mathbf{M}\begin{bmatrix}[c||c] \mathbf{x}^w_O & 1 \end{bmatrix} \\
    % \mbox{\ \  and \ \ }
    &\lambda_1 \mathbf{K}^{-1}\begin{bmatrix}[c||c] \mathbf{x}^p_1 & 1 \end{bmatrix} = \mathbf{M}\begin{bmatrix}[c||c] \mathbf{x}^w_1 & 1 \end{bmatrix}
\end{split}
\end{equation}
From these two we get 6 equations to solve for five unknowns: three for $\mathbf{T}$ and one for $\lambda_0$ and $\lambda_1$ each.

\subsection{Obtaining the Annotations}
\label{sec:annotations}

For these estimations, we need three annotations in an image as discussed in \secref{sec:estim-relat-posit}: (1) world origin and axes, (2) three orthogonal vanishing points, and (3) one known dimension. We now describe how we obtain these annotations
using deep neural networks (DNNs), and \textbf{\textit{when that is infeasible}}, crowd-tasking on Amazon Mechanical Turk.

%% : One known dimension (length, height, or width) of some object in
%%   the camera view.
%% \item The world origin (and associated axes) from which the world
%%   coordinate location of one other point can be calculated using the
%%   known dimension.
%% \item Vanishing directions that are consistent with the world axes
%%   directions.
%% \end{enumerate}
% \begin{figure*}
%     \centering
%     \subfloat[]{\label{fig:turker:1}\includegraphics[width=0.45\linewidth]{images/turker_page.png}}\qquad \qquad
%     \subfloat[]{\label{fig:turker:2}\includegraphics[width=0.35\linewidth]{images/vanish_point_turker.eps}}
%     \caption{(a) Web UI for Annotation Crowdsourcing. (b) Sample Annotation Collected. Red, purple, and yellow lines correspond to x-axis, y-axis, and z-axis in the world coordinate frame of reference.}
%     \label{fig:turker}
% \end{figure*}

\begin{figure*}
    \centering
    \subfloat[]{\label{fig:turker:1}\includegraphics[width=0.30\linewidth]{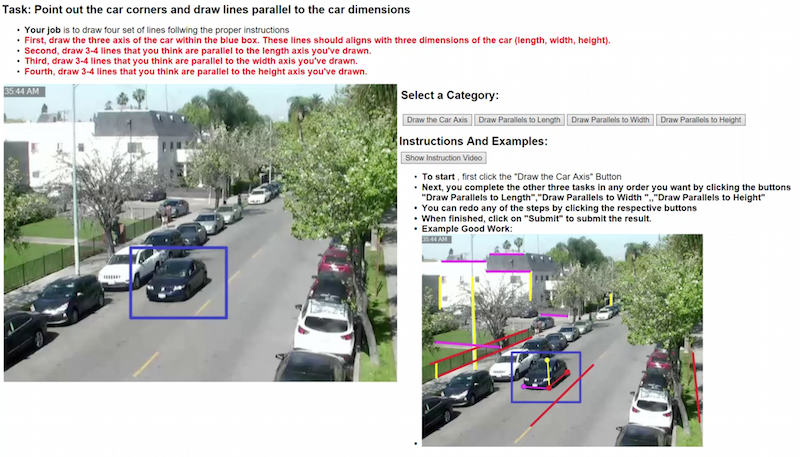}}
    \subfloat[]{\label{fig:turker:2}\includegraphics[width=0.23\linewidth]{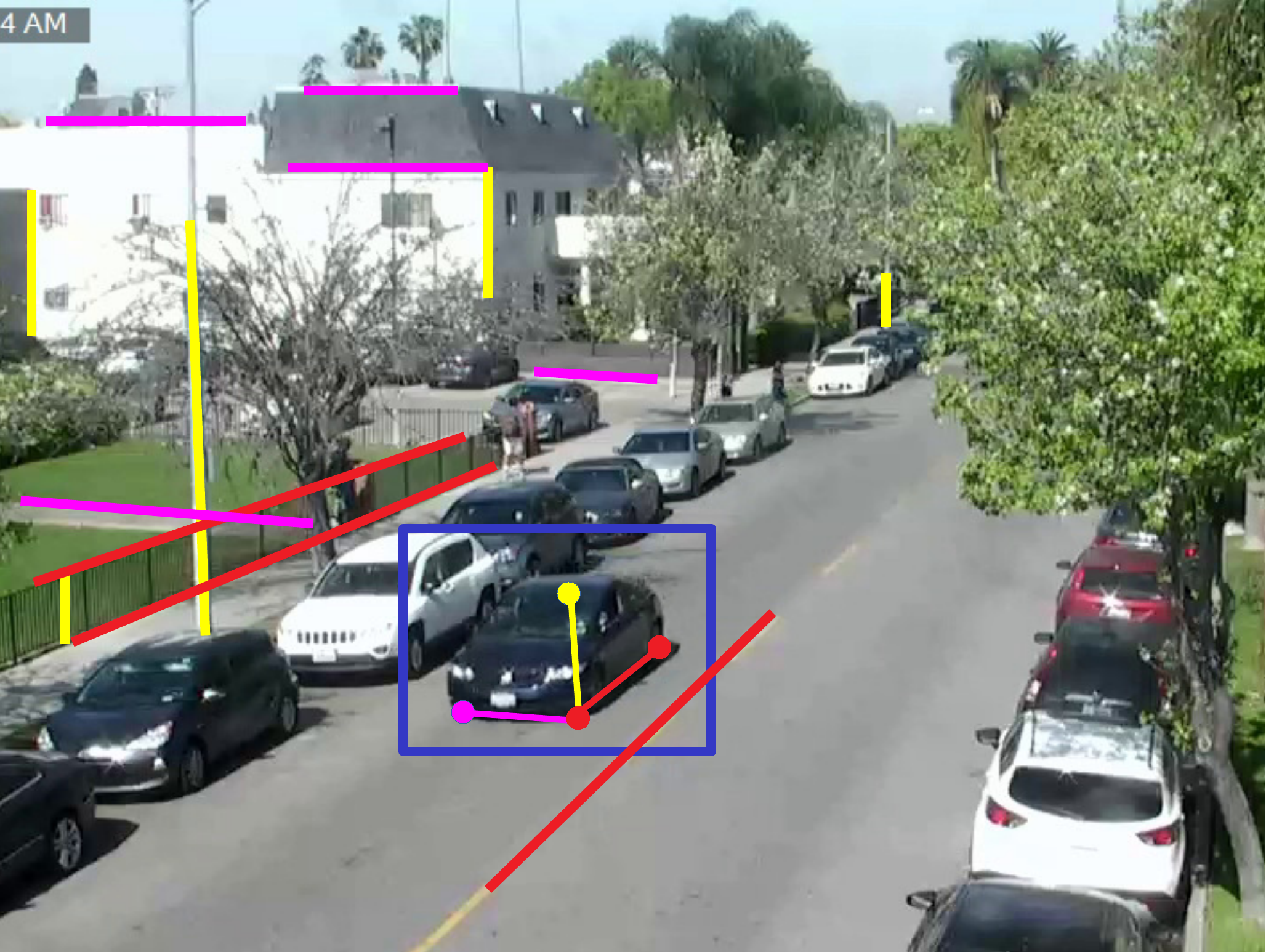}} \quad
    \subfloat[]{\label{fig:app_illustretion}\includegraphics[width=0.45\linewidth]{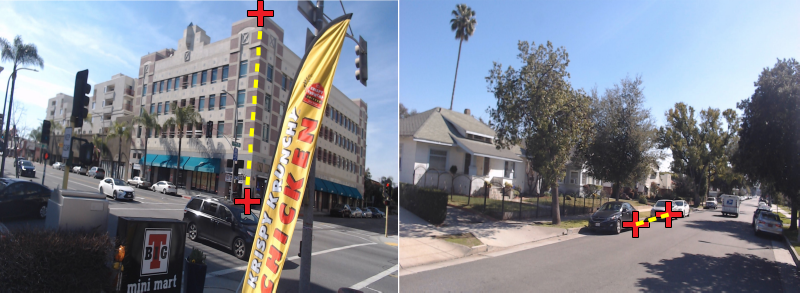}}
    \caption{(a) Web UI for annotation crowdsourcing. (b) Sample annotation collected. Red, purple, and yellow lines correspond to x-axis, y-axis, and z-axis in the world coordinate frame of reference. (c) Illustration of building height (left) and parking space annotation (right).}
    \label{fig:turker}
\end{figure*}

% \begin{figure}[!ht]
%     \centering
%     \includegraphics[width=\linewidth]{images/app_illustration.png}
%     \caption{Illustration of Building Height (Left) and Parking Space Annotation (Right)}
%     \label{fig:app_illustretion}
% \end{figure}

%% In this section, we detail the proposed system for Relative Position
%% estimation. As explained in the Section~\ref{sec:rel_loc}, we need
%% three sets of information in order to estimate the relative position
%% from geometry. Thus, our proposed system for relative localization
%% consists of four building blocks as illustrated in
%% Fig.~\ref{fig:rel_sys_illus}: (1) Car Detection and Dimension
%% Estimation (CaDDE) (2) Car Axis Annotation (CAAn), (3) Orthogonal
%% Vanishing Point Detection (OViD) and (4) Relative Pose Estimation
%% (ReaPE). Next, we explain each of the modules along with how they
%% interconnect. The input to the system is an image sampled from the
%% camera feed.

\parab{One known dimension.} For this, we leverage the observation that street-surveillance cameras usually have one or more four wheeled vehicles (cars, trucks \etc) in their view. We first use an object detector to detect the vehicle, and then describe how we can estimate one of the dimensions of the vehicle.

\parae{Detecting vehicles.} For this, we use a standard object detector DNN, SSD~\cite{liu2016ssd}, which outputs bounding boxes as well as associated labels. However, for the next step in the annotation (described below), we need to find a vehicle (among those detected by SSD) whose three dimensions (length, width, and height) are fully visible. To do this, we leverage the observation that larger bounding boxes that are closer to the image center are likely to have the property we need. So, we pick that bounding box whose ratio of area to distance from the box center to the image center is highest.

\parae{Estimating car dimensions.} For this, we use a pre-trained neural network~\cite{mousavian20173d} that, given a car image, outputs the car dimensions. Because this neural network is sometimes inaccurate, we also experiment with using fixed estimates for these dimensions (\secref{sec:experiment}) based on the observation that most sedans have roughly the same size to within a meter or so.

\parab{Obtaining world origin and axes.} For this, we use humans to annotate the four points shown in \figref{fig:caraxis}: the origin, and one point along each of the dimensions of the car. We use an image annotation crowd-sourcing service~\cite{Satyam} for this purpose. This service enables users to programmatically submit tasks to annotate images (\eg draw bounding boxes, track them, segment objects). It then automates the task of sending annotation requests to workers on Amazon Mechanical Turk (AMT) (called \textit{Turkers}) and of collecting and curating results. We extended the service to incorporate our annotations. In our approach, the service sends each image, with the vehicle bounding box identified in the previous step, to multiple turkers. We describe below how we aggregate results from these Turkers.

Till date, we know of no \emph{reliable} automated method for obtaining these annotations, which we have left it to future work.

\parab{Obtaining vanishing points.} To determine a vanishing point, we need to find two lines in the image that are parallel in the real world (\eg the sides of a rectangular building, the sides of a car). For our estimation to work (\secref{sec:estim-relat-posit}), we need these lines to align with the axes identified in the previous step.\footnote{As a matter of detail, the annotated vehicle need not be parallel to the street, but the annotations for the vanishing points must align with the car.}

For this, we evaluate two approaches. In the first, we use the Canny edge detector to find all edgelets (small edges) in the image. Then we use the car axis directions and the RANSAC method~\cite{Fischler:1981:RSC:358669.358692} to filter out the edgelets that are not consistent with the world axis direction then estimate the three vanishing points using~\cite{6385802}.

Because this method is sometimes unable to find three vanishing points, we use crowd-sourcing to obtain these annotations. In this method, the Turker is asked to draw 3-4 lines that are parallel to each of the car axis. We use the same service described above~\cite{Satyam}, and have extended the service to obtain vanishing point annotations.

\figref{fig:turker:1} shows the webpage the service presents to each Turker. Each Turker annotates the car axes and three sets of parallel lines. \figref{fig:turker:2} shows a sample result from a Turker, which depicts three sets of mutually orthogonal lines pertaining to the vanishing points.

%\pradipta{Ramesh: we need to add something about the minimality of annotation}

%% The next module in the
%% pipeline is tasked with annotating the car axis. Assuming that the car
%% length, width and height represent the X, Y, Z directions of the world
%% coordinate, we are interested in finding the pixel locations of four
%% points $\mathbf{X}^p_0$, $\mathbf{X}^p_1$, $\mathbf{X}^p_2$,
%% $\mathbf{X}^p_3$ corresponding to the four points on the car
%% $\mathbf{X}^w_0 = [0,0,0]^T$, $\mathbf{X}^w_1 = [L,0,0]^T$,
%% $\mathbf{X}^w_2 = [0,w,0]^T$, $\mathbf{X}^w_3=[0,0,H]^T$,
%% respectively. This is illustrated in Fig.~\ref{fig:caraxis}.

%% To this end, we employ Amazon Mechanical Turk based crowd-sourcing
%% where humans are tasked to visually point out the four points of
%% interest in each query image. An automated method is also possible by
%% employing image segmentation on an sequence of images from the video.
%% However, we opt for a crowd-sourcing system as human have in-build
%% perception of directions and it fulfills our goal of showcasing the
%% proof-of-concept of the proposed pipeline. Development of methods to
%% autonomously detect these points based on image-processing and
%% deep-learning is left as a future work.

\

\parab{Putting it all together: Robustly estimating the camera's relative position.} Our annotations are \textit{minimal}: we \textbf{\textit{get just the information needed by the underlying mathematical framework}}. From these annotations, we can estimate the intrinsic and extrinsic matrices, then obtain the camera's relative position as discussed above. In practice, we may need to aggregate multiple \textit{candidate locations} in order to get a robust position estimate. Since untrained Turkers may produce inaccurate results, we generate candidate locations from each worker's annotations. Furthermore, since the car dimension estimator is sometimes erroneous, we obtain two different estimates of the translation matrix: one from the car's length and the other from the car's width. Thus, for each Turker, we get two candidate locations. We then cluster these locations by distance, and estimate the camera position by the centroid of the largest cluster.

\subsection{Applications}
\label{s:applications}

Beyond estimating the relative position of the camera, we can estimate, under some conditions, the world coordinate $\mathbf{x}^w$ corresponding to \textit{any} given pixel coordinate $\mathbf{x}^p$. To do this, we can use the pixel projection equation (\eqnref{eq:pixprojection}): $\lambda \begin{bmatrix}[c||c]{\mathbf{x}^p} & 1\end{bmatrix} = \mathbf{K}\mathbf{M}
\begin{bmatrix}[c||c] {\mathbf{x}^w} & 1\end{bmatrix}$. However, in this system, there are three equations with four unknowns $\lambda$ and the three world coordinates.

Under some conditions, we can use this capability to \textbf{\textit{employ the camera as a virtual sensor}}.

%% to estimate \emph{dimensions} of a feature on a street, which leads to several interesting applications.

%% \hang{This is fascinating. One possible use case for this, is to build
%%   an accurate correspondence between image pixel and lidar output,
%%   without any prior knowledge of the relative location between the two
%%   devices. On the other hand, this correspondence can help calibrate
%%   the relative location estimate between the two devices. }

\parae{Virtual scale.} Just as scales can measure dimensions, the camera's position, together with human annotations, can be used to estimate lengths or distances. If a human could annotate a pixel on the ground plane (as shown in \figref{fig:app_illustretion} (right)), then there are only 3 unknowns (since the z-axis value is zero, assuming that the world coordinate system is on the ground plane as is the case with our annotations), so it is possible to obtain the world coordinate of this pixel using \eqnref{eq:pixprojection}. We show that, if a human annotates two pixels that demarcate the edges of some feature on the ground plane (\eg a parking space or a road divider) we can estimate the length of this feature.

\parae{Virtual clinometer.}  A clinometer measures building height. Our approach can synthesize a virtual clinometer from the camera, as follows.
% \textbf{\textit{Building height.}}
If a human can annotate a pixel $\mathbf{x}^p_1$ where a building's face meets the ground, and a pixel $\mathbf{x}^p_2$ at the top of the building vertically above $\mathbf{x}^p_1$ (as illustrated in \figref{fig:app_illustretion} (left)), we can estimate the building height. This is because, since $\mathbf{x}^p_1$ is on the ground, there are only 3 unknowns in estimating the world coordinate of $\mathbf{x}^p_1$. Once we know $\mathbf{x}^p_1$'s world position, we can determine $\mathbf{x}^p_2$'s since its $x$ and $y$ coordinates are the same as $\mathbf{x}^p_1$'s. The Euclidean distance between these two points is the height of the building.

\parae{Virtual radar.} A radar measures vehicle speeds. From a camera's consecutive frames, we can obtain a cheap, but approximate, vehicle speed estimator. To do this, we use a feature based tracker such as KLT~\cite{klt}, which tracks image features across multiple frames. To estimate the speed, we estimate the distance traveled by a feature belonging to a car ($\delta$) in successive frames and use the camera's frame rate ($fps$) to estimate the speed ($\delta * fps *3.6$~kmph).

\section{Absolute Localization}
\label{sec:abs_loc}

Obtaining the absolute location of the camera (in terms of its latitude and longitude), given its relative location, is a significant challenge. As with relative location, we need additional contextual information to determine absolute location. In this section, we begin by describing how to determine absolute location using minimal additional information, then discuss how to obtain this additional information in practice.

%% This section shows how we can use the relative position estimation
%% of the Camera (obtained by the techniques discussed in \secref{sec:rel_loc})
%% to estimate the absolute GPS coordinate i.e., latitude-longitude (lan-lon) of the camera. 
%% First, we explain theory underlying this relative to absolute location
%% transformation process which we have developed. This explanation identifies 
%% the requirement of pixel location to lat-lon location mapping of at least
%% one point in the image to obtain a set of possible absolute location of the camera. 
%% Next, we detail how such mapping can be obtained by focusing on two particular
%% real-worlds scenarios. 

\subsection{Estimating Absolute Position}
\label{sec:estim-absol-posit}

\parab{Overview:} The intrinsic and extrinsic matrices (\secref{sec:estim-relat-posit}) give us the world coordinates of the camera. Let $\mathbf{x}^p_1$ be the pixel coordinate of a ground plane point. Using the pixel projection equation (\eqnref{eq:pixprojection}), we can obtain that point's global coordinate, and, from that the relative position of the camera \textit{with respect to that ground plane point}. Then, if we knew the absolute position of pixel $\mathbf{x}^p_1$, we can just use it to find the absolute position of the camera. To do this, we need to: (a) obtain the absolute position of the image pixel, and (b) the mapping between our world coordinates and the geodetic (latitude/longitude) coordinate system.

%% This is
%% difficult, for two reasons:
%% \begin{enumerate}[itemsep=0pt,leftmargin=*]
%% \item It is a very challenging task to obtain the GPS coordinate of an
%%   image pixel just from the image.
%% \item We do not know the mapping between our Global Coordinate system
%%   to the GPS coordinate system. Without this mapping information, the
%%   relative position estimations can not be directly applied. More
%%   specifically: How distance in meters maps to difference in lat-lon
%%   coordinate? What the X, Y axes direction of Global coordinates
%%   correspond to on the earth surface?
%% \end{enumerate}

%% It turns
%% out that knowing the GPS coordinates of a pixel, $\mathbf{x}^p_1$,
%% itself cannot get us the GPS coordinate of the camera. We are still
%% missing information on the mapping between the Global coordinate and
%% the GPS coordinate system. It can only help us narrow the search space
%% for the camera location to a circle around the pixel GPS coordinate,
%% $(lat_1, lon_1)$, with a radius of
%% $d^w_{rel,1} = \sqrt{(\delta x^w_1)^2 + (\delta y^w_1)^2}$. Again to
%% get the circle in terms of lat-lon we need to convert the distance to
%% the lat-lon difference.
% \begin{figure}
%     \centering
%     \includegraphics[width=0.8\linewidth]{images/abs_dir_illustration.eps}
%     \caption{Illustration of Different Components of the Absolute Position Estimation. $(\delta_x^w, \delta_y^w)$ is the estimated position of the camera relative to the landmark pixel.}
%     \label{fig:abs_ill}
% \end{figure}

\begin{figure*}
    \centering
    \subfloat[]{\label{fig:abs_ill}\includegraphics[width=0.3\linewidth]{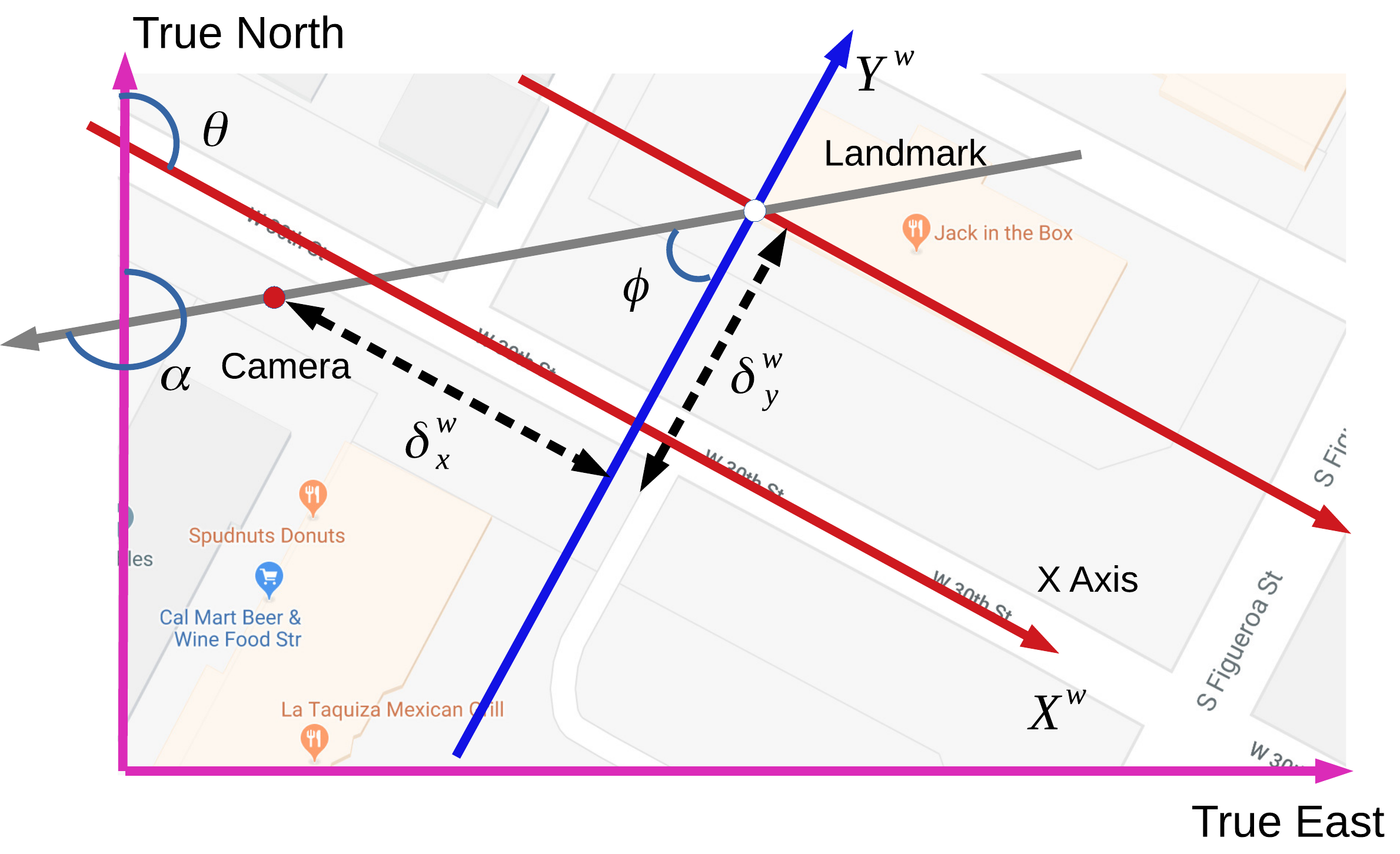}}
    \subfloat[]{\label{fig:fourcandidate}\includegraphics[width=0.18\linewidth]{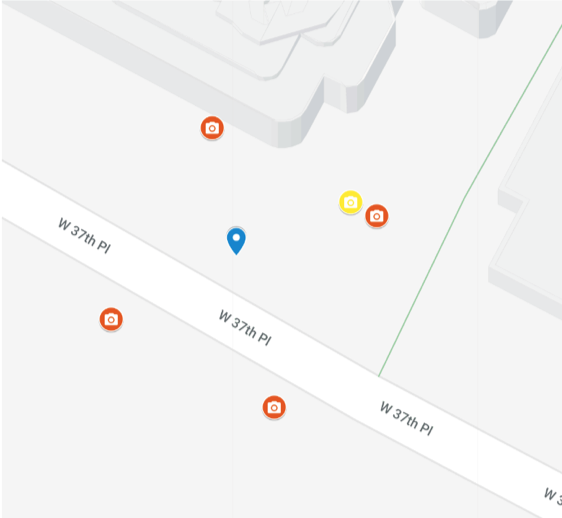}} \quad
     \subfloat[]{\label{fig:two_point_ill}\includegraphics[width=0.49\linewidth]{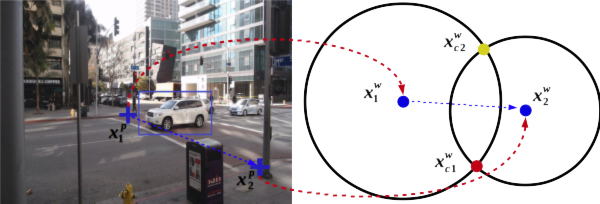}}

    \caption{(a) Illustration of different components of the absolute position estimation. $(\delta_x^w, \delta_y^w)$ is the estimated position of the camera relative to the landmark pixel. (b) Candidate locations based on one reference point. The yellow marker is the actual camera location. Red markers are the estimated camera location candidates. (c) Estimating a unique camera location using two pixel-to-GPS mappings.}
    % \label{fig:abs_ill}
\end{figure*}
% \begin{figure}[!ht]
%     \centering
%     \includegraphics[width=0.5\linewidth]{images/four_candidate.png}
%     \caption{Candidate Locations based on One Reference Point. Yellow marker is the actual camera location, Red markers are the estimated camera location candidates.}
%     \label{fig:fourcandidate}
% \end{figure}
% \begin{figure}
%     \centering
%     \includegraphics[width=\linewidth]{images/two_point_illustration.png}
%     \caption{Estimating a Unique Camera Location using Two Pixel-to-GPS Mapping.}
%     \label{fig:two_point_ill}
% \end{figure}

\parab{Given the absolute position of a pixel.} Let $(lat_{1},lon_{1})$ be the known absolute position of the pixel (\secref{sec:pixtogps} discusses how we can obtain this), and let $d$ be the distance and $\alpha$ the bearing from that point to the camera. Then, prior work~\cite{aviat} provides the following approximation for small $d$ (less than 1-2~kms, which holds for camera ranges) for the absolute position $(lat_{c},lon_{c})$ of the camera:
\begin{equation}
  \begin{split}
    lat_{c} &= lat_{1} + (d \cdot \cos{\alpha}) / 111111\\
    lon_{c} &= lon_{1} + (d \cdot \sin{\alpha}) / (111111\cdot \cos(lat_{1}))
  \end{split}
  \label{eqn:gps}
\end{equation}
where the constant 111111 is the distance on the Earth's surface corresponding to 1 degree change in latitude and $\alpha$ is the bearing (clockwise) towards the true north.

%% where $\alpha \in (-180, 180]$ is the bearing of the camera position
%% from the pixel's GPS coordinate with respect to true north direction
%% with the positive direction towards the True East, illustrated in
%% \figref{fig:abs_ill}. You can also can notice from \eqnref{eqn:gps}
%% that if the bearing $\alpha$ is known, we can uniquely solve for the
%% lat-lon for the camera, ($lat_c, lon_c$).

%% The
%% latitude-longitude coordinates are the polar coordinate of a location
%% with respect to earth's center. Prior work on different earth model
%% approximation~\cite{aviat} can relates latitude-longitude difference
%% to distance over earth's surface. Such models considers the earth
%% surface to be flat for the purpose of distance measurement between two
%% points if they are located less than couple Kilometers apart. Since
%% the typical range of surveillance cameras are much less that 1 Km, we
%% can approximate the region as flat. In a spherical earth model, 1
%% degrees of latitude implies 111111 meters of surface displacement
%% whereas the displacement for one degree of longitude corresponds to
%% $\cos{latitude}$ times the displacement due to 1 degree of latitude.
%% With this information, we can map the circular region to a set of
%% lat-lon coordinates as follows:

\parae{Estimating $\alpha$:} Public APIs for Google Maps and Open Street Maps provide the angle (clockwise) of the street $\theta$ relative to the true north. Now, suppose one of the world axes (say the x-axis) aligns with the direction of the street. Then, knowing the relative position of the camera, we can estimate the bearing of the camera, $\phi$, relative to the pixel point in the world coordinates. From this, we can estimate $\alpha$, the bearing of the camera relative to the pixel point in the geodetic system to be $\alpha = 90 + \phi + \theta$ (\figref{fig:abs_ill}) where all the values are in degrees.

Unfortunately, there is still an important source of ambiguity: from the image, we cannot know which direction of the street the camera is facing. For example, if the street runs from southwest to northeast, the street's bearing in that direction is 45 degrees, but in the opposite direction is 225 degrees. From this, we get two possible values for $\alpha$. This arises because we don't know how the world x and y axes relate to the geoidal directions. For each choice of bearing, there are two choices for the y-axis direction. This leads to four candidate locations for the camera when using (\eqnref{eqn:gps}), as shown in \figref{fig:fourcandidate}.

\parab{Given the absolute position of two pixels.} In this case, it becomes easier to estimate the absolute position of the camera. From the pixel coordinates of the pixels, we can estimate the world coordinate locations of the pixels, and their distances ($d_1$ and $d_2$) to the camera. The camera must lie at one of the two points at which the circles centered at each pixel, with radius $d_1$ and $d_2$ respectively, intersect. We can find the absolute positions of these two points (using \eqnref{eqn:gps}). There still remains ambiguity in selecting one location.

A human annotator or a simple comparison of the pixel coordinates of the reference pixel can eliminate this ambiguity. In \figref{fig:two_point_ill}, $\mathbf{x}^p_1$ and $\mathbf{x}^p_2$ are the two pixels and $\mathbf{x}^w_{c1}$ and $\mathbf{x}^w_{c2}$ are the candidate locations. By examining the relative ordering of $\mathbf{x}^p_1$ and $\mathbf{x}^p_2$ when scanning the camera image from left to right, we can determine whether $\mathbf{x}^w_{c1}$ or $\mathbf{x}^w_{c2}$ is the correct absolute position. In our example, if $\mathbf{x}^p_1$ is to the left of $\mathbf{x}^p_2$, $\mathbf{x}^w_{c1}$ is the correct position, $\mathbf{x}^w_{c2}$ otherwise.

\subsection{Obtaining Absolute Position of  Pixels}
\label{sec:pixtogps}

To obtain the absolute position, \textbf{\textit{we must have some coarse grain information about the camera's location}}.

\parab{Road Intersection.} If we know the intersection near which the camera is located, and if at least one of the corners of that intersection is visible on the camera, we can narrow down the absolute location of the camera to a handful of candidates. For this, we rely on human annotation.

%% Let us consider a practical scenario where
%% the metadata from the public website tells you that the camera is
%% located near a particular road intersection and at least one of the
%% intersection corners is visible in the video feed. This is very common
%% for the traffic cameras. Now the questions is: how do we use this
%% information to get the the pixel to GPS mapping for the camera?

\begin{figure}
    \centering
    \includegraphics[width=\linewidth]{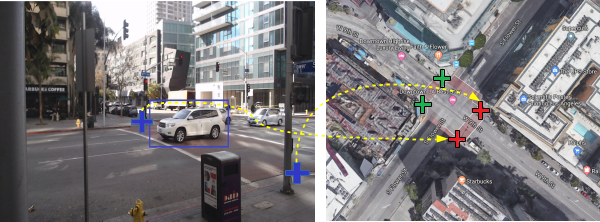}
    \caption{Two-corner-based absolute position estimation. The top two pairs of candidate positions are illustrated as in red and green, respectively.}
    \label{fig:twocorner}
\end{figure}

We present both the image and the satellite view of the respective road intersection to a human annotator as shown in \figref{fig:twocorner} and ask him/her to mark all the corners in both images in a \textit{clockwise} manner. We then have an ordered list of visible corners in the image, and, using a map service, we can get the GPS coordinates of each intersection point.

This leads to another ambiguity: it is impossible for a human operator (without additional information) to identify which corner in the image corresponds to which corner in the satellite view. This ambiguity results in candidate positions whose number depends on the number of corners visible in the camera image.

%% The next obvious question is: how can we associate the corners in the
%% image with the correct physical corners of the intersection if no
%% other information is available? WLOG, let us assume that there are
%% four corners on the intersection as four is the most common corner
%% count in intersections around the World. Next, we will discuss four
%% different cases and our respective methods of matching the corners.

% \parae{One visible corner:} 
\parae{One visible corner:} If only one corner is visible, it can map to one of the 4 corners in the satellite image. For each of these points, we get 4 candidate locations (\secref{sec:estim-absol-posit}), resulting in 16 candidates for the absolute location. We have assumed that the intersection has only four corners; our analysis can be extended to more complex intersections.

\parae{Two visible corners:} 
% \textbf{\textit{Two visible corners:}}
With two visible (adjacent) corners, there are 8 possible mappings between these and the four corners in the satellite view. This results in 8 candidate solutions (for each choice, we can exactly pinpoint the location, \secref{sec:estim-absol-posit}). We can narrow these down further as follows. We first estimate the distance between the two visible corners using the method discussed in \secref{sec:annotations} for finding dimensions in an image. From the absolute positions of the corners obtained from the satellite view, we can also find the distances between the intersection corner points. We can match the image distance to the closest satellite view distance. This works well when two streets of different widths intersect. In this case, we can reduce the number of candidates to two (\figref{fig:twocorner}) by symmetry.

%% However, we can get some
%% additional information from our system. From the corner pixel
%% annotations and applying \eqnref{eq:pixprojection} with the estimated
%% camera parameters, we can get the position of both corners relative to
%% the camera. Using this information, we can estimate the relative
%% position between the corner points and, subsequently, the length of
%% the line segment joining the corners. On the other hand, from the
%% Google map or Open-Street-Map, we can get the actual distance between
%% the annotated corner points. Next, we can simple match the distances
%% and take the side with a length closest to the estimated length.
%% However, it turns out that due to symmetric nature of opposite sides
%% of a typical road intersection and error in the length estimation, we
%% do not always get the correct combination by using the closest match.
%% Nonetheless, we observed that the correct combination is always within
%% the top two combinations. Thus, we take the top two corner pairs to
%% get a set of 4 possible locations. One instance of such matching is
%% illustrated in \figref{fig:twocorner}.

\parae{More than 2 visible corners.} With 3 or 4 visible corners, we can use similar matching techniques: in this case, we would match the sequence of distances between successive corners (assuming that the corner annotations are clockwise). Even so, however, because intersections are symmetric, there are always at least two candidate matches for any sequence of corner distances, so we cannot completely eliminate the ambiguity.

\parab{Landmark Building.} If the camera is near a known landmark building (such as a restaurant or a retail store) whose GPS location is available using a map API, and the landmark is also visible in the camera, then we can get some candidate absolute positions in at least two ways: asking the human to annotate a \textit{building corner} or the mid-point of the \textit{building face} and assuming that annotated pixel's GPS location is the same as the building's. In either case, similar ambiguities exist as with using road corners; we omit the details of this estimation for brevity, but present results in \secref{sec:experiment} to show how well this approach works.

\subsection{Applications}
\label{s:applications-1}

Using these annotations, we can estimate the camera's absolute position as described in \secref{sec:estim-absol-posit}. From this, we can estimate the absolute position of any ground pixel in the image.

%% \added{Using our proposed system and two landmark pixel coordinates, we can estimate the distance of any ground pixel from the camera and the landmarks. Then using the estimated location of the camera, we can triangulate the GPS coordinate of any pixel in the image.}

\parab{Virtual guide.} Using this, we can track a person appearing in an image whose GPS trace is approximately known, a capability we call a \textit{virtual guide}. This can provide navigation guidance for visually-challenged users. Suppose such a user has a cellphone that continuously updates a cloud service with its own location. The cloud service can, using a nearby camera, estimate where the person is on the camera (because it can obtain the pixel position of the user's GPS location), then alert the user to obstacles on the way (\eg fire hydrants, objects on the sidewalk etc.). In practice, because GPS locations can be inaccurate, the cloud service will need to (a) use a DNN to detect bounding boxes for the person, and \textit{track} their absolute positions over time (by converting successive pixel positions to absolute positions) and (b) \textit{match} the tracks with the GPS tracks. To do this tracking robustly in the presence of multiple users, we use a technique from~\citet{tar}.

\section{Evaluation}
\label{sec:experiment}

In this section, we evaluate relative and absolute position accuracy using a custom dataset containing 214 images, and compare its performance against  PoseNet~\cite{Kendall_2015_ICCV}, a state-of-the-art Deep Neural Network for camera localization.
% \pradipta{There was one concern that why we do not compare with other DNN technqiues that are similar to Posenet but newer. we didn't do it because they are of same class and used on same dataset. Not sure how to address that here.}

\subsection{Methodology}
\label{sec:methodology}

\begin{figure}
    \centering
       \includegraphics[width=0.95\columnwidth]{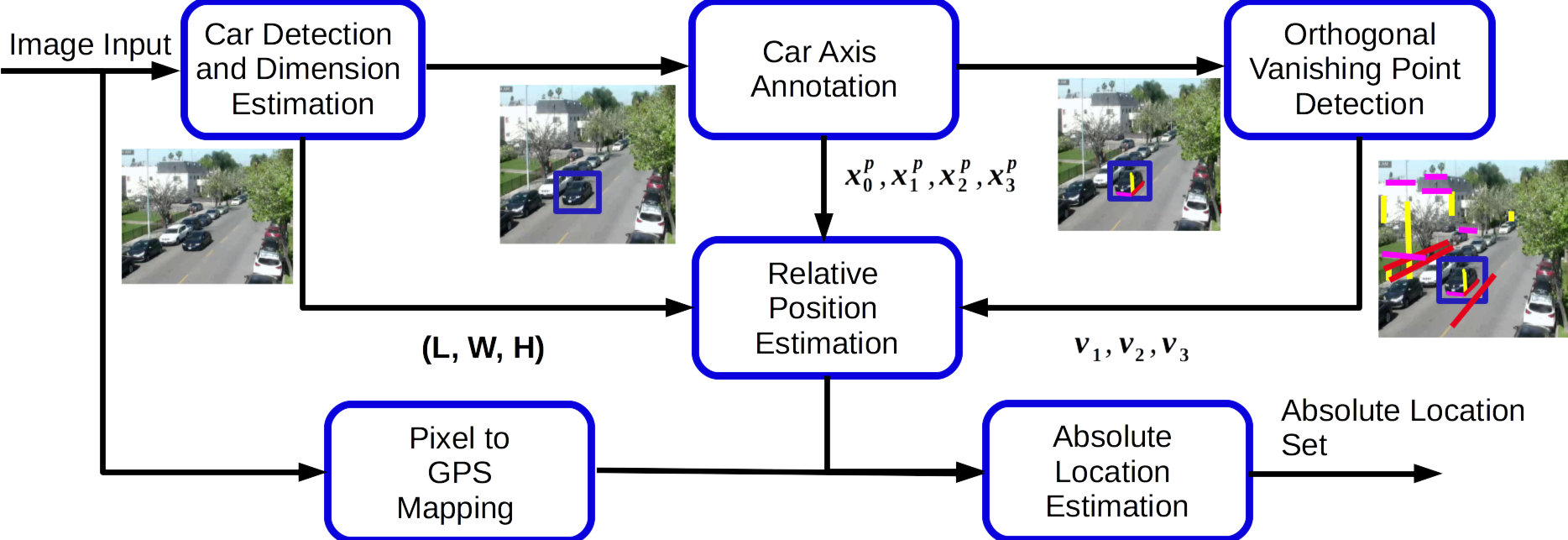}
        \caption{CamLoc workflow}
    \label{fig:camloc_system}
\end{figure}

\parab{Implementation.} We have instantiated our algorithms into an end to end system, called CamLoc, which has six components as shown in \figref{fig:camloc_system}. The \emph{Car Detection and Dimension Estimation} component takes the camera image as input, outputs the 2D bounding of the most-visible car along with its three dimensions (L, H, W). For estimating dimensions, CamLoc uses standard sedan dimensions. \textit{Car Axis Annotation} uses crowdsourcing to obtain the world origin and the three world axes (\figref{fig:caraxis}). \textit{Orthogonal Vanishing Point Detection} extracts the three orthogonal vanishing directions using crowdsourced annotations. \textit{Relative Position Estimation} uses the techniques described in \secref{sec:rel_loc} to estimate the relative position of the camera, while \textit{Pixel to GPS Mapping} obtains pixel annotations using crowdsourcing (\secref{sec:pixtogps}). Finally, \textit{Absolute Location Estimation} outputs the candidate absolute positions using techniques described in \secref{sec:abs_loc}. Overall, CamLoc requires 1,680 lines of code in Python.

% \hang{Is LoC needed?}

%% In this section, we details our implementation of the camera
%% localization system instance, that we refer to as CamLoc, which
%% employs the concepts discussed in \secref{sec:rel_loc} and
%% \secref{sec:abs_loc}. The CamLoc system has 6 component as illustrated
%% in \figref{fig:camloc_system}: (1) Car Detection and Dimension
%% Estimation (2) Car Axis Annotation, (3) Orthogonal Vanishing Point
%% Detection, (4) Relative Position Estimation, (5) Pixel to GPS Mapping,
%% (6) Absolute Location Estimation. Next, we briefly explain each of the
%% modules along with how they interconnect. The input to the system is
%% an image sampled from the camera feed.

%% This module perform two sequential tasks: \emph{Car Detection and Car
%%   Dimension estimation.} For the car detection, we employ a very well
%% known pre-trained object detection tool called
%% SSD~\cite{liu2016ssd}~\footnote{\url{https://github.com/balancap/SSD-Tensorflow}}.
%% For the car dimension estimation, we use a standard predetermined
%% dimensions of a Sedan (detailed in \secref{sec:rel_loc}). We also have
%% a provision of using a pre-trained DNN for dimension estimation. For
%% the purpose of comparative experiments presented in
%% \secref{sec:experiment}, we will refer to the system with DNN as
%% \emph{``CamLoc-NN''} to differentiate from the normal implementation
%% of CamLoc.

%% This module outputs the three orthogonal vanishing points using the
%% Amazon Mechanical Turk based annotations detailed in
%% \secref{sec:annotations}. For the purpose of comparison in
%% \secref{sec:experiment}, we also have a provision of using
%% automated edge detection based RANSAC method for three orthogonal
%% vanishing point detection~\cite{Fischler:1981:RSC:358669.358692}.
%% To differentiate from the default implementation of CamLoc, we will
%% refer the the CamLoc system with automated vanishing point
%% detection as \emph{``CamLoc-Auto.''}

%% To get the pixel to GPS mapping in CamLoc, we rely fully on human
%% turkers/annotators. Human turkers are asked to perform three types
%% of annotations: (1) Exact mapping of two pixel points to the GPS
%% coordinate by showing them the query image and the a small region
%% around the camera's true location, (2) Annotations of the visible
%% corners in the image and the corners of the known intersection in
%% Clockwise order for the known road intersection scenario detailed
%% in \secref{sec:pixtogps}, and (3) Annotation of landmark midpoint
%% and corners for the known landmark scenario detailed in

%% We also have an instance of CamLoc where all the annotations are
%% done by a single experience human annotator to test CamLoc system's
%% performance with a single good annotation. We will denote this
%% instance of CamLoc as \emph{``CamLoc-Expert-Anno''}.
\begin{figure*}
    \centering
    \subfloat[CDF of the Position Estimation Error]{\label{fig:rel_error:1} \includegraphics[width=0.32\linewidth]{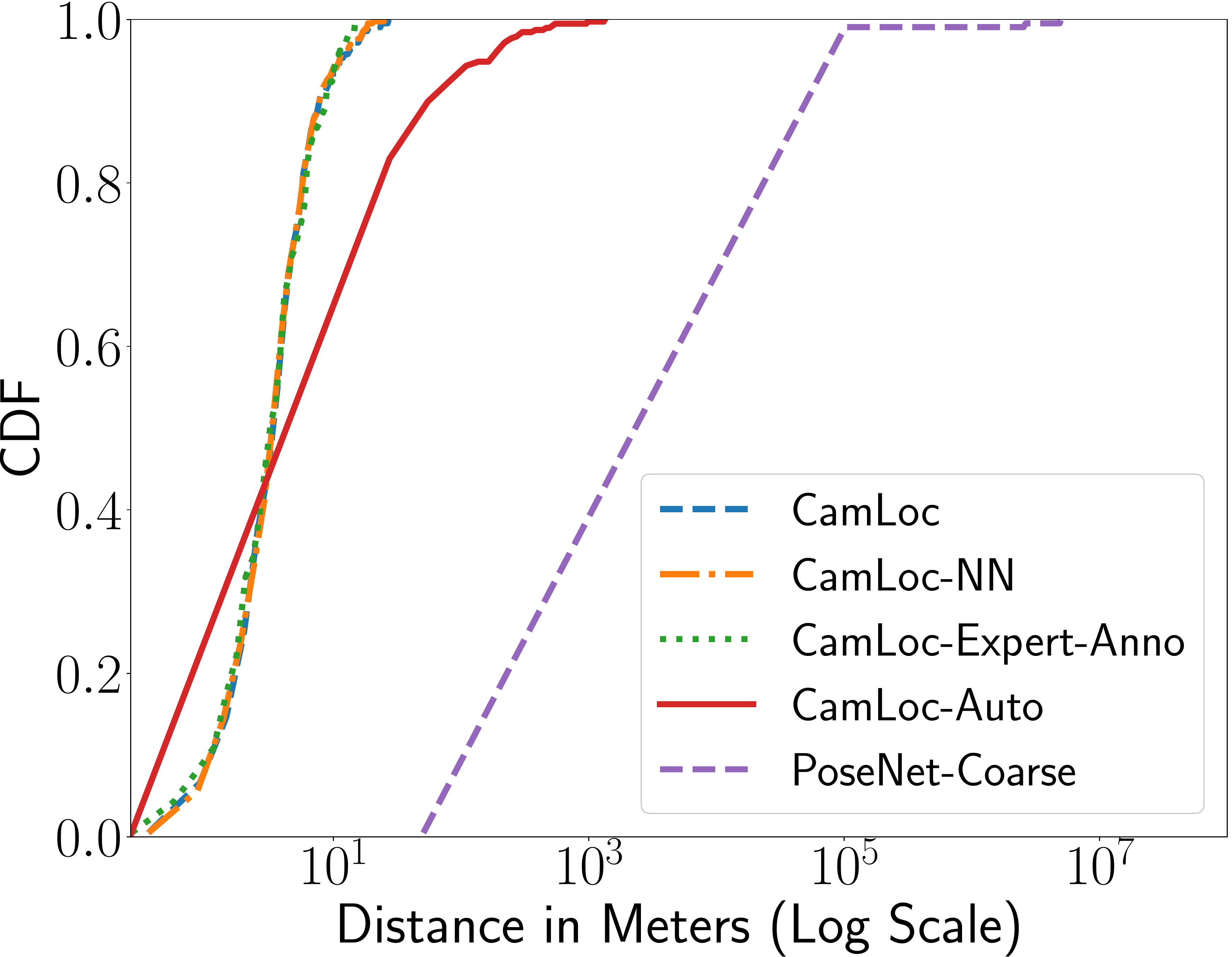}}\,
    \subfloat[CDF of the Height Estimation Error]{\label{fig:rel_error:2}\includegraphics[width=0.32\linewidth]{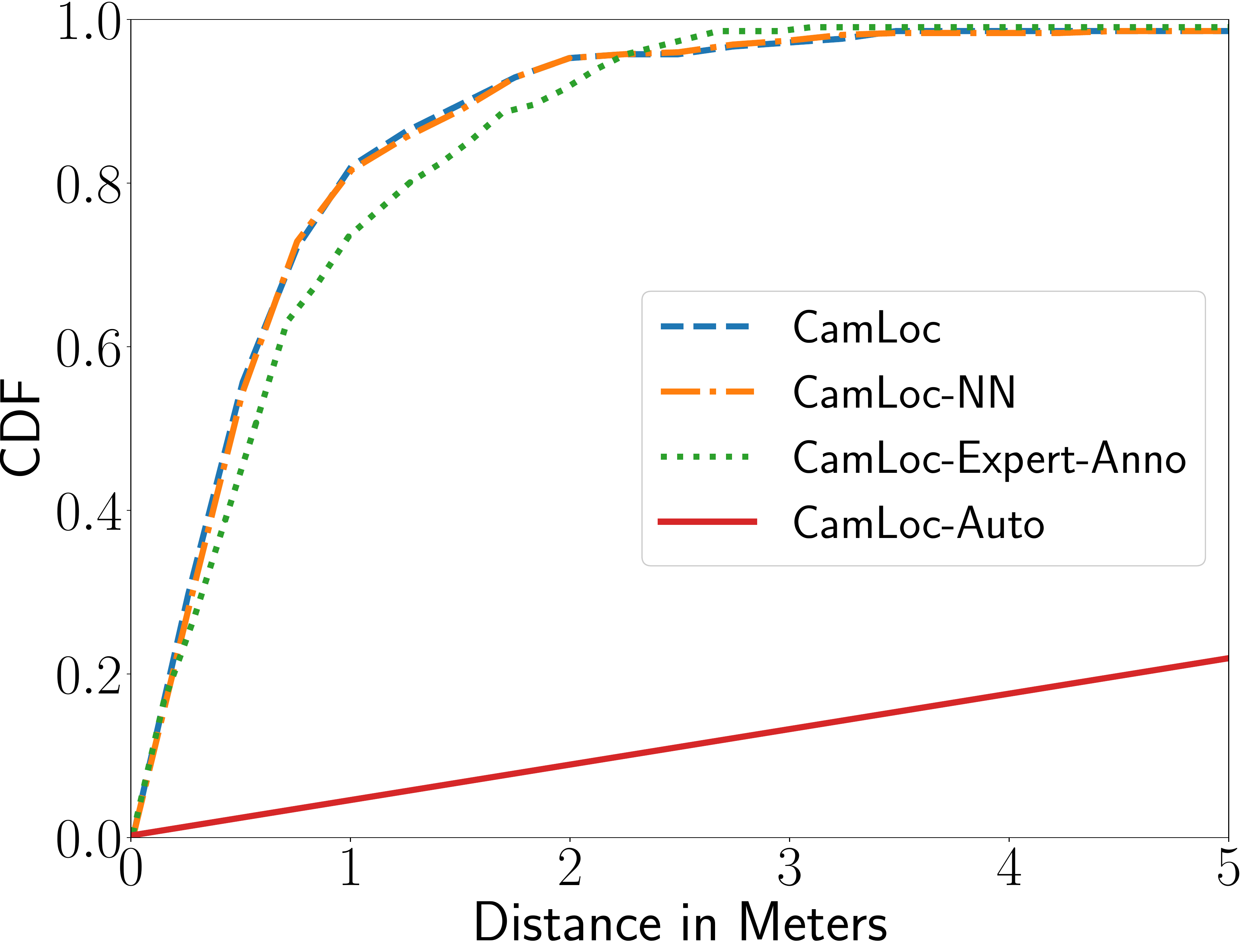}}
    \subfloat[]{\label{fig:bad_image} \includegraphics[width=0.33\linewidth]{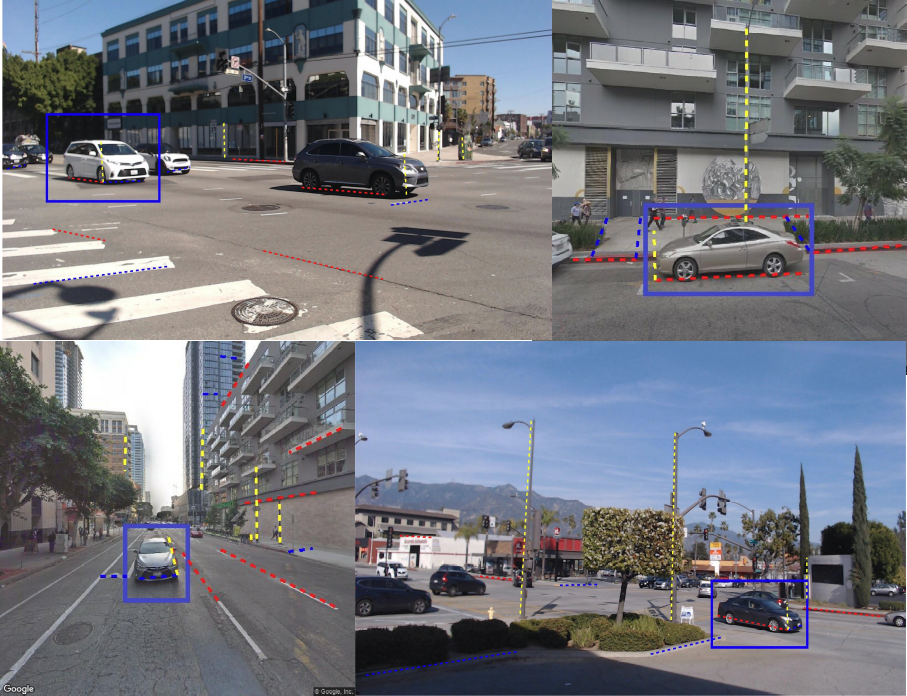}}
    \caption{(a, b) Relative position error analysis, (c) Four sample images that are hard to annotate. The top-left image contains a corner where intersecting roads are not perpendicular. For the rest three images, one of the dimension is not too apparent at first look.}
    \label{fig:rel_error}
\end{figure*}

\parab{Dataset.} CamLoc localizes a camera from a single image taken by the camera. We were unable to find an open-source image database with ground-truth locations and having the diversity of camera poses, camera parameters, and image quality in surveillance cameras. Google Street View provides some diversity and ground truth, but all images are at street level from cameras with identical properties. To have more diversity, we use a dataset of 214 images in which roughly $50\%$ of the images are from Google Street Views. Another $40\%$ of the images we collected using smartphone cameras from roughly 80 different combinations of locations, pose and elevations around a large city in North America. The remaining $10\%$ are snapshots from different public online cameras such as EarthCam and three campus surveillance cameras in the same city.\footnote{We plan to make the dataset publicly available upon publication.} Our dataset also has, by design, diversity in image resolution; despite this, as we show below, CamLoc is able to estimate camera position well. 

\parab{Ground Truth.} For our evaluations, we annotated, using a tool we developed, the ground truth GPS location of the camera and the selected car in each image along with the relative location. Our annotation tool combines GPS measurement, satellite views, and human-in-the-loop to output the ground truth. We also use this annotation tool to obtain ground truth for our applications.

%% Moreover, for the evaluation of the two application of related
%% localization: parking space estimation and building height
%% estimation; we leverage similar satellite view based annotation
%% along with open-street-map data.

\parab{Comparison.} We compare the performance of CamLoc with PoseNet~\cite{Kendall_2015_ICCV}. To this end, \textbf{\textit{we trained PoseNet on the Street View dataset of a large metropolis containing 368,000 images}} totaling up to 24 GB. Further, to compare how PoseNet performs when we have more context for the absolute location and can narrow the search space to a smaller region, \textit{\textbf{we retrain PoseNet on a much smaller dataset that contains 74,000 street view images from our campus neighborhood.}} We  refer to these two separate models as \emph{PoseNet-Coarse} and \emph{PoseNet-Fine}, respectively.\footnote{Recent research has improved PoseNet; we describe why we do not compare against these in \secref{sec:related_work}.}

% \pradipta{To add difference Between PoseNet Dataset and Our Dataset.}

To illustrate the impact of our design choices for some of the components, we also compare CamLoc with the following alternatives: \textit{CamLoc-NN} uses a pre-trained neural net to detect vehicle dimensions; \textit{CamLoc-Auto} uses an automated approach to detect vanishing directions~\cite{6385802}; and \textit{CamLoc-Expert-Anno} uses annotations by an expert to understand the efficacy of crowdsourcing.

\parab{Experimental setup.} The experiments run on an Intel Core i7 8700 CPU @ 3.20GHz$\times$12 machine with GeForce GTX 1050 Ti GPU.

\subsection{Relative Positioning Accuracy}
\label{sec:relative}

In this section, we evaluate the relative localization performance of CamLoc system. We compare CamLoc with different variants of CamLoc and \textit{PoseNet-Coarse}. PoseNet outputs the absolute position of the camera. For the purpose of comparison, we convert this to the relative position with respect to the world origin using the camera projection equation (\eqnref{eq:camprojection}).

\parab{Position Estimation Error.} In \figref{fig:rel_error:1}, we compare the error in the relative position estimation in meters between the estimated location and the ground truth location. The relative position error in CamLoc is less than 5 meters for $\approx 80\%$ of the images and less than 10 meters for 95\% of the images. A careful analysis of the images reveals that errors larger than 5~m occur when one dimension of the car/environment is not visible or unclear. This results in the human annotator not being able to draw parallel lines correctly, so CamLoc is unable to determine three vanishing points. We show four examples of such images in \figref{fig:bad_image}. The other prevalent reason for the high error is that Turkers annotate the wrong car corner as the world origin (we ask them to annotate the corner closest to the camera). This can lead to an error up to the length of the car $\approx 4.5 m$.

\parae{CamLoc outperforms PoseNet.}
Figure~\ref{fig:rel_error:1} clearly demonstrates that CamLoc outperforms PoseNet \textit{\textbf{by nearly 2 orders of magnitude}}. When trained over a corpus of images in a large area, PoseNet loses the ability to precisely match images and determine pose. Two factors account for the discrepancy between the sub-meter localization error presented in its original evaluation~\cite{Kendall_2015_ICCV}. First, the \textit{\textbf{testing and training images}} in PoseNet's original evaluation \textit{\textbf{use the same camera}} at a relatively consistent height. In contrast, our training images use Google Streetview images taken at street level but our test images are from multiple surveillance cameras, smartphones, and webcams with different camera properties from a wide range of heights and orientations. PoseNet is unable to generalize well to these. Second, the evaluation in~\cite{Kendall_2015_ICCV} \textit{\textbf{uses images taken from a few hundred meters stretch of road, much smaller in scope than our city-wide training dataset}} (tens of square kms). At these larger scales, Posenet is unable to effectively extract features unique to specific camera poses,  leading to inferior performance. By contrast, CamLoc uses precise geometric techniques, together with human annotations tailored towards the specific image, in order to determine relative position.

%% \emph{In summary, Posenet cannot generalize to large areas: it performs well if (1) the images are all from a similar camera at a similar angle and (2) there exists a common prominent feature in all the images. The diversity in location and orientation in our test dataset makes it hard for PoseNet to perform well.}

\parae{CamLoc is better than other variants.}
CamLoc outperforms CamLoc-Auto in the tail by an order of magnitude. Automated vanishing point detection can fail to find three vanishing directions in situations where human annotators are able to find these (we discuss this in more detail below). CamLoc-Expert-Anno, with annotations from an expert, performs almost the same as CamLoc which incorporates annotations from multiple untrained turkers. This is likely due to the ``wisdom of the crowds'' phenomenon; the aggregated results from multiple untrained workers have been repeatedly shown to match ground truth well. Finally, CamLoc-NN, which uses a neural net to estimate car dimensions, does not perform appreciably better than CamLoc; this suggests that it might be sufficient to use standard dimensions for this component.

\begin{figure*}[!ht]
    \centering
    \subfloat[]{\label{fig:abs_error}\includegraphics[width=0.33\linewidth]{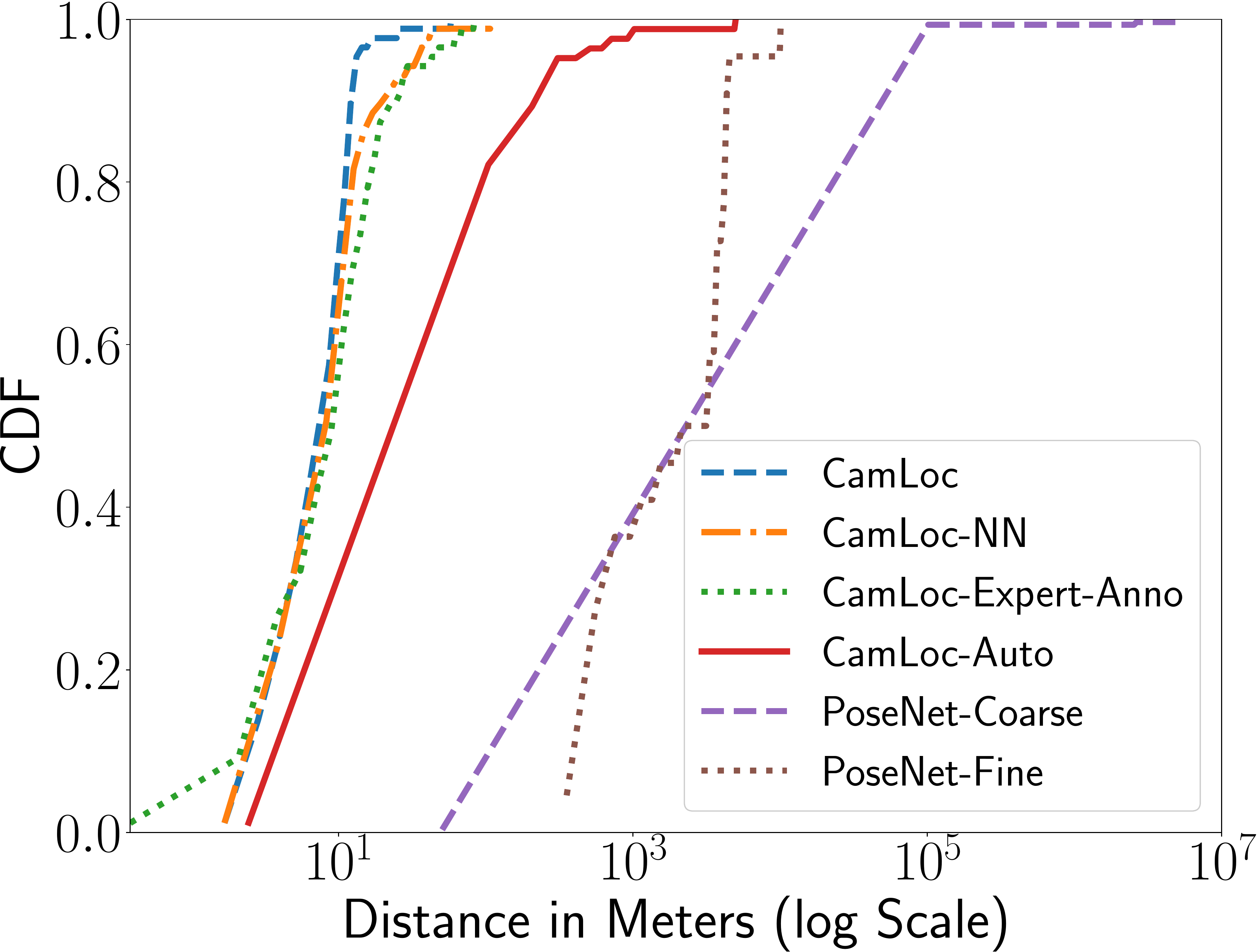}}
    \subfloat[]{\label{fig:abs_error_sc}\includegraphics[width=0.33\linewidth]{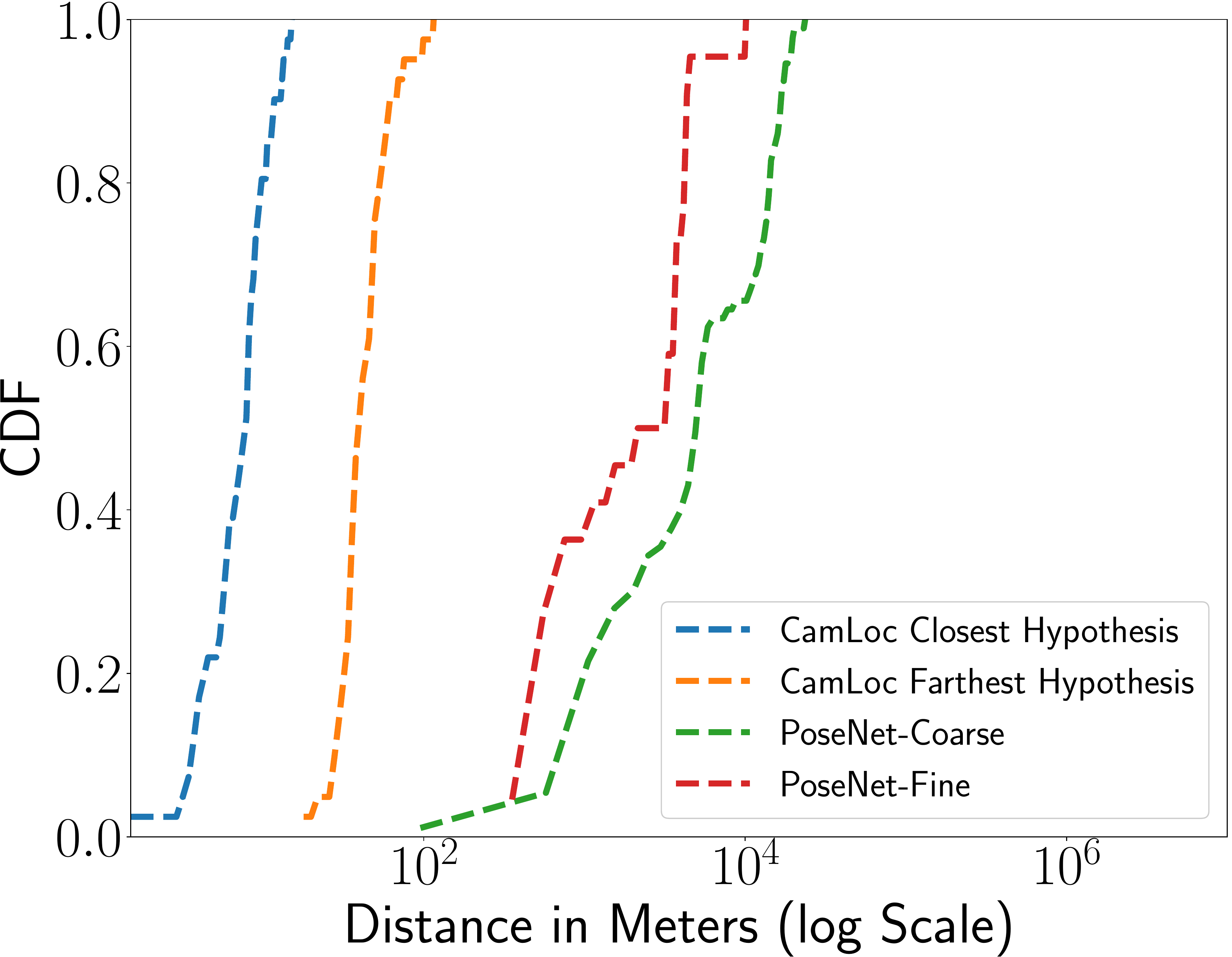}}
    \subfloat[]{ \label{fig:abs_error_lm}\includegraphics[width=0.33\linewidth]{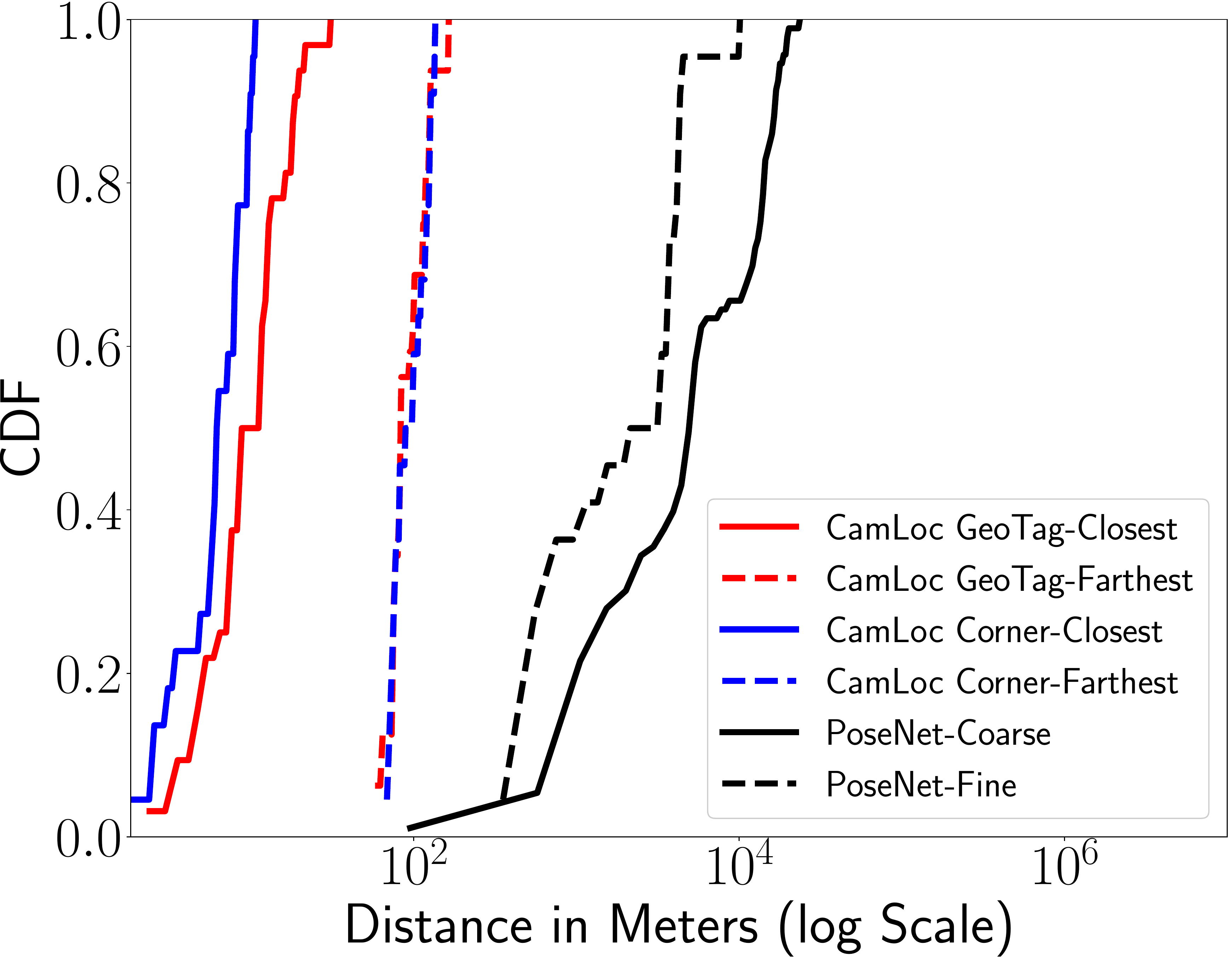}}
    \caption{Absolute position error analysis for (a) two pixel-to-GPS mapping based estimation, (b) road corners based estimation, and (c) landmark building based estimation.}
    % \label{fig:abs_error}
\end{figure*}

%% We can
%% further observe that the CamLoc-NN which uses DNN for estimating the
%% car dimension does not have any performance edge over the fixed
%% standard dimension based estimation in CamLoc. It suggests that we do
%% not need to employ DNN for the car dimension estimation.

\parab{Camera Height Estimation Error.} From its relative position estimates, we explore how well CamLoc can estimate camera height (\figref{fig:rel_error:2}). In this experiment, we do not compare CamLoc with PoseNet, since the latter is trained on Street View images all taken from the same height.

\parae{CamLoc estimates camera height better than alternatives.}
\figref{fig:rel_error:2} shows that CamLoc with human-in-the-loop can achieve less than 1~m error in height estimation for 80\% of the test images and less than 3~m error in 95\% of the cases. CamLoc-Auto performs the worst. As with position error, errors higher than 1~m occur when at least one dimension of the car/environment is not visible/prominent, as illustrated in \figref{fig:bad_image}.

\parae{Why CamLoc-Auto performs poorly.} While automated detection of the most promising vanishing point can be very accurate, detecting the second and third orthogonal vanishing point is very challenging with automated line detection and grouping of parallel lines. Often the automated approach outputs three vanishing points that are not orthogonal at all. In cases where the system is able to detect a set of orthogonal vanishing points, it most often does not align well with the car axes and therefore can not leverage the car dimensions to estimate the camera location. This is exactly what led us to design crowd-sourced annotations for vanishing points.

%On the other hand CamLoc does not rely on any such training and takes each image as a isolated input.

%% In summary, CamLoc system outperforms all other variants as well as PoseNet-Coarse.

% \begin{figure}
%     \centering
%     \includegraphics[width=\linewidth]{images/bad_turker.png}
%     \caption{Four sample images that are hard to annotate. The top-left image contains a corner where intersecting roads are not perpendicular. For the rest three images, one of the dimension is not too apparent at first look.}
%     \label{fig:bad_image}
% \end{figure}

\subsection{Absolute Positioning Accuracy}
\label{sec:absol-local}

In this section, we analyze the error in absolute position estimation. For this, we perform three sets of experiments as described below.

%% In the second, we analyze the performance of
%% CamLoc when only closest road intersection information is known and at
%% least one of the intersections is visible in the image. Finally, we
%% analyze CamLoc performance when at least one Landmark is visible in
%% the image.

\parab{Two Known Points.} 
% \noindent\textbf{Two Known Points.}
In the first set of experiments, we get the pixel to GPS mapping for two pixels by leveraging a combination of landmark location, road intersections, and different features on the road such as a left-turn arrow or stop sign text. From these, we extract the absolute positions of two pixels and compare CamLoc against four variants and PoseNet-Coarse.

In this experiment, we also ask: would PoseNet perform better if its models were geographically specialized? That is, if we knew the precise area covered by a set of cameras, and we trained PoseNet with images only from that area, would it perform better? For this, we compare CamLoc against PoseNet-Fine (\secref{sec:methodology}).

\parae{CamLoc outperforms other alternatives.}
\figref{fig:abs_error} clearly shows that CamLoc with human annotations can achieve less than 12 meters errors for 95\% of the images. This performance is consistent with the relative position error performance of CamLoc. Both variants of PoseNet perform really poorly on the test images with errors that are one or two orders of magnitude worse than CamLoc. While PoseNet-Fine performs slightly better than PoseNet-Coarse, it is still an order of magnitude worse than CamLoc. The performance of CamLoc-Auto is also much worse that CamLoc again due to either not being able to detect three orthogonal vanishing points or detecting a set of orthogonal vanishing points that are not aligned with the car axes.

For relative positioning, CamLoc was comparable to CamLoc-NN and Camloc-Expert-Anno. However, for absolute positioning, it is better than these two alternatives. CamLoc is better than CamLoc-Expert-Anno because the wisdom of the crowds has a stronger effect in this case, where the turkers are able to collectively annotate pixel to GPS mappings better. CamLoc outperforms CamLoc-NN since the latter inconsistently and inaccurately estimates car dimensions across multiple runs. In summary, for the case with known mapping of two ground pixels with their GPS location, CamLoc can achieve localization accuracy in the order of a few meters, comparable to GPS error in congested city environments.

We do not report height estimation error since that is the same both with absolute and relative positions.

% \begin{figure}
%     \centering
%   \includegraphics[width=0.9\linewidth]{results/Distance_Error_CDF_abs_road_corners.pdf}
%     \caption{Absolute Position Error Analysis for Road Corners Based Estimation}
%     \label{fig:abs_error_sc}
% \end{figure}
% \begin{figure}[!ht]
%     \centering
%     \includegraphics[width=0.95\linewidth]{results/Distance_Error_CDF_abs_LM.pdf}
%     \caption{Absolute Position Error Analysis for Landmark Building Based Estimation}
%     \label{fig:abs_error_lm}
% \end{figure}

\parab{Road Corners.}
We now consider a more restrictive case, when the only available information is that the camera is near a known intersection and the corresponding road corners are visible in the image. For this set of experiments, we handpicked a set of 50 representative images where the road intersections are visible. In this setting, recall that CamLoc outputs multiple candidate locations.

\figref{fig:abs_error_sc} plots the CDF of the distance to the closest candidate location output of CamLoc as well as the distance to the furthest candidate location from the ground truth location. The closest candidate location is within 10~m of the actual location for almost more than 80\% of the test cases and within 15~m for 100\% of the test cases while the furthest candidate is sometimes 100~m away. Even the furthest candidate location is orders of magnitude closer to the ground truth than the PoseNet-Coarse and PoseNet-Fine.

% \begin{figure}
%     \centering
%   \includegraphics[width=0.9\linewidth]{results/Distance_Error_CDF_abs_road_corners.pdf}
%     \caption{Absolute Position Error Analysis for Road Corners Based Estimation}
%     \label{fig:abs_error_sc}
% \end{figure}

\parab{Landmark Building.} In this section, we present CamLoc performance for the case when a landmark building is visible in the image. As discussed in \secref{sec:pixtogps}, CamLoc can use two annotations: landmark midpoint annotation on the most prominent visible face, or annotations on visible corners. In our experiments, we use both of these methods and obtain a set of candidate locations from them.

\figref{fig:abs_error_lm} plots the minimum and maximum distance of the candidate locations from the ground truth location for both cases. With corner annotations, CamLoc output always contains a candidate location which is less than 10~m away from the camera true position. This number compares well with the absolute localization performance of CamLoc with two known pixel positions.

On the other hand, for the case of mid-point annotations that use the geo-location of the building, the closest candidate is less than 15 meters for approximate 80\% of the images and less than $25$ meters for 95\% of the test cases. This difference arises from the fact that the annotated midpoint may not always correspond to the extracted geo-location. Our compensation for this difference helps, but cannot ensure high accuracy for all images. In both cases, the furthest candidate location is within 100~m of the actual location which is still orders of magnitude better than the performance of the PoseNet-Fine and PoseNet-Coarse.

% \begin{figure}[!ht]
%     \centering
%     \includegraphics[width=0.95\linewidth]{results/Distance_Error_CDF_abs_LM.pdf}
%     \caption{Absolute Position Error Analysis for Landmark Building Based Estimation}
%     \label{fig:abs_error_lm}
% \end{figure}

% \hang{We came to the conclusion that farthest results should be kept. But, can we convert 100m error to percentage here? Since the farther the building is out there, the harder it is to get the same low absolute error in meters.}

\subsection{Applications}
\label{sec:relat-locat-appl}

CamLoc's positioning abilities allows cameras to act as virtual sensors of different types. We now evaluate how well these virtual sensors work.

\begin{figure}[!ht]
    \centering
    \includegraphics[width=\linewidth]{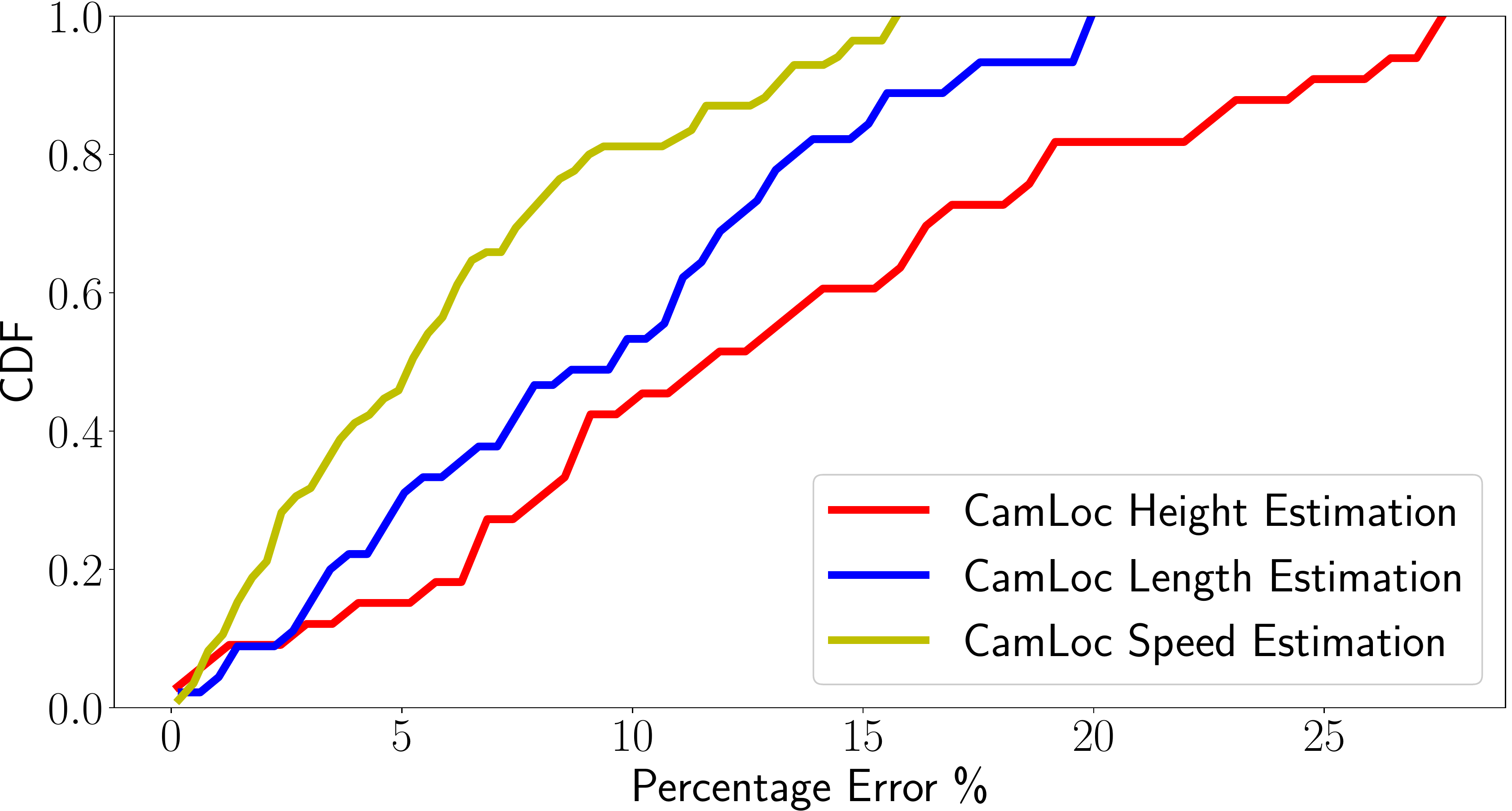}
    \caption{Percentage error for virtual scale, virtual clinometer, and virtual radar.}
    \label{fig:application_Error}
\end{figure}

\parab{Virtual Scale and Virtual Clinometer.} We select 50 images in our set with marked parking spaces or road dividers (\textbf{\textit{camera as virtual scale}}, \secref{s:applications}), and annotate two pixels corresponding to lengths ranging from 10~m to 80~m (we obtained the ground truth for these lengths by pinpointing positions on the satellite view). For  building height estimation (\textbf{\textit{camera as virtual clinometer}}, \secref{s:applications}), we handpicked a set of 50 images with visible full buildings of height ranging from 3~m to 55~m (the height ground truth is from Open Street Maps~\cite{osm}). In \figref{fig:application_Error}, we plot the ratio (expressed as a percentage) of the error to the actual dimension. The virtual scale's error is less than 20\% for all of the test cases. On the other hand, the virtual clinometer's error is less than 25\% with 90 percentile and less than 30\% with 100 percentile. The larger tail for building height estimation comes from smaller buildings: CamLoc has an absolute error of 1-2~m, which translates to high percentage error in some of the small buildings (our smallest is 3~m).

\parab{Virtual Radar.} For estimating vehicle speed (\textbf{\textit{camera as virtual radar}}, \secref{s:applications}), we selected a 10~min worth of video from three of the campus cameras with 100 vehicles with speeds ranging from 10km/hr to 100km/hr (we obtained the groundtruth by pinpointing the entry and exit positions of the vehicles on the satellite view).
%% In \figref{fig:application_Error}, we plot the ratio (expressed as a % percentage) of the error to the speed.
For speed estimation, the error (\figref{fig:application_Error}) is less than 15\% overall and less than 10\% at the 80th percentile. The larger error corresponds to estimating the speed of very fast moving vehicles.

\parab{Virtual guide.} To demonstrate the use of the camera as a virtual guide (\secref{s:applications-1}), we set up a surveillance camera along a street and asked four volunteers to randomly walk in the camera's view with an Android phone to log their GPS locations. We applied the method detailed in \secref{sec:estim-absol-posit} to match the GPS traces with the persons detected on the camera. Our four GPS traces are all correctly matched after a maximum of 2 seconds of bootstrapping time.

\parab{Summary.} While the virtual sensors are less accurate than the physical ones, they can provide \textit{triage}: for instance, the virtual radar can be used to detect streets with persistent speeders to deploy speed traps, and the virtual clinometer can be used to estimate the approximate height of a skyline for aerial advertising.

\subsection{Resource Usage}
\label{sec:micro-benchm-analys}

In this section, we analyze CamLoc in terms of time and resources (compute and human) required for the crucial components.

\parab{Annotation Effort.} Our dataset includes in total 214 images. We collected 3053 annotations, so on average 14.3 annotations per image, among which approximate 85\% are valid annotations, and 15\% were due to spammers on AMT. The median time to finish the annotation for one image is 130.95 seconds or 146.95 person-hours across our entire dataset. For a single camera, annotation is a one-time task, so the required annotation effort required is reasonable.

\parab{Run-time of CamLoc with available annotations.} We performed a small experiment to estimate the time required to get the results for each image, once we have the annotation. CamLoc takes $\approx 0.41$ seconds per image and CamLoc-NN takes $\approx 0.42$ seconds per image.

\section{Related Work}
\label{sec:related_work}

\parab{Camera localization.} Most of the work in this area focuses on estimating the 6 degrees of freedom (6-DOF) of a camera that can be subdivided into three broad categories.

\parae{Feature Matching based Image Localization:} This class of related work mainly focus on identifying the place/view that the image corresponds to.  A large class of approaches~\cite{Bansal_2014_CVPR,7410667,7924572,ArmaganHRL17}  use different image descriptors such as SIFT~\cite{Lowe2004} features, ORB~\cite{ORB2011} features, bag of visual words~\cite{cula2001compact} or spatial features~\cite{philbin2007object} to compare the query image with a large database of geo-tagged annotated images. Some~\cite{Zamir2010StreeView,Chen2011Landmark} also compare the image features with the features in Google StreetView dataset. For example, \citet{storefronts2015} classify different types of storefronts captured in Google Street View. However, such techniques suffer from two inherent problems: (1) the location of the image is not same as the location of the camera, and (2) the database of images is often sparse and might not contain any matching images~\cite{valentin2015exploiting}. In addition, for surveillance camera images, the quality of the images and the angle of the cameras make it even harder to properly identify and match the image descriptors. Given matched image, the relative camera pose between the query image and the matched image can provide an estimate of the camera's position This requires the correspondence of at least 5 pixel coordinates in both images~\cite{5Points2004, Zhang06}. For these reasons, CamLoc uses geometric techniques.

%% We do not employ such system due to its really bad performance with
%% our dataset images.

\parae{3D model-based localization:} The second class of techniques employs a pre-estimated 3D representation of the region of interest for a 2D-3D matching of the features extracted from the query image~\cite{sattler2015hyperpoints,sattler2017efficient}. Structure from motion methods can provide such a representation~\cite{koenderink1991affine}. To generate a set of 2D-3D correspondences for such camera relocalization, \citet{Shotton_2013_CVPR} propose a regression forests method using RGB-D images; \citet{Cavallari2017CVPR} extend this by adapting a pre-trained forest to a new scene. However, this class of methods heavily relies on the availability and quality of a 3D map of the environment which is expensive, time-consuming, and often not available in urban environments. In contrast, CamLoc does not require any 3D map of the regions or RGB-D query image for localization.

\parae{Neural network based camera pose estimation:} PoseNet~\cite{Kendall_2015_ICCV} utilizes a convolutional neural network to regress the camera pose (6-DOF) from a single RGB image. It maps monocular images to a high dimensional space that is linear in pose. This representation allows for full 6-DOF camera pose using regression. \citet{7487679} extends  PoseNet to incorporate a Bayesian model that determines the localization's uncertainty. NetVLAD~\cite{NetVLAD2018}, besides the standard CNN layers, adds a generalized Vector of Locally Aggregated Descriptors (VLAD)~\cite{VLAD2010} layer, which is a popular descriptor pooling method for image retrieval and classification. \citet{Kendall_2017_CVPR} showed that PoseNet can improve performance by taking into account geometric loss functions and scene re-projection error. \citet{li:hal-01867143} present a new angle-based re-projection loss function to solve the problem of the original reprojection loss, enabling neural network training without careful initialization. However, as shown in \secref{sec:experiment}, PoseNet does not generalize well to the diverse set of query images in our dataset. VLocNet~\cite{VLocNet2018} and later VLocNet++~\cite{VLocNet++2018} use consecutive monocular images and apply multi-task learning (MTL) for learning semantics, 6DoF pose regression, and visual odometry. The performance of all these approaches were mainly tested on the Microsoft 7-Scenes Dataset~\cite{7-scene} which is a small indoor dataset and the Cambridge Dataset~\cite{cambrid-data} which is an outdoor dataset consisting of consecutive images from couple blocks taken from a single camera. Like PoseNet, all subsequent extensions are likely to perform poorly due to the lack of generalization. In contrast, the proposed CamLoc system does not rely on any training while generalizing well to images with significant perspective diversity. (As an aside, only PoseNet code is available, so it is the only alternative we evaluate).

% \citet{Laskar_2017_ICCV} presented a CNN for querying a database of geo-tagged images of known poses to estimate the relative pose between the query and the database images. Our preliminary experiment with such a system shows that such systems also fail to properly extract and match feature for images from the surveillance camers. 
%VLocNet++ outperforms PoseNet on 7 Scenes dataset.

\parab{Vanishing Point Detection.} There exists a large literature on vanishing point detection in the vision community~\cite{kovsecka2002video,lezama2014finding,Vanishing2003}. However, accurate vanishing point detection involves grouping of multiple parallel lines in the physical world in order to estimate the respective vanishing point. Detection of such parallel lines in images and grouping them can yield a large number of vanishing points that includes many false positives. Thus, vanishing point detection remains a challenging problem~\cite{simon2016simple} to date. Moreover, we need three mutually orthogonal vanishing points instead of just one vanishing point. While it may be easy to accurately detect 1 to 2 orthogonal vanishing points, in our experience it is much harder and error-prone to detect 3 orthogonal vanishing points in a real world setting. Some approaches use a RANSAC based approach for detecting three orthogonal vanishing points either by relying on known focal length~\cite{6385802} or a known principal point~\cite{wildenauer2012robust}. However, the outputs of these approaches often may not align properly with the three car dimensions selected by our annotators. Our crowdsourcing based vanishing point detection leverages human perception of parallel lines for robust estimation of the vanishing points.

% \textbf{Geopositioning} For absolute position system, \citet{Imbriaco2019geopositioning} compare a query image against a panoramic image database. The comparison uses an aggregation of memory vectors from global image descriptors based on convolutional features to improve querying the database. 
% \citet{GeographicalWebCam2014} globally localizes IP webcam with 24 hours image sequence. They segment the sky and classify images as day or night by comparing the mean luminance. Then, sunrise and sunset are estimated. Lastly, the latitude and longitude are computed.
% Im2GPS~\cite{IM2GPS2008} and PlaNet~\cite{PlaNet2016} try to determine where a given photo was taken on the Earth's surface. Instead of an image retrieval approach, they modeled it as a classification problem. They split the surface of Earth into a grid of geographic cells. They train a convolutional neural network with millions of photos. They require GPS-tagged images.

%Descriptor based approach can be used for finding pose parameters but it requires more than one image for descriptors comparison. 

\parab{Other Related Work.} The work of \citet{Xu2014} is complementary to our work that uses computer vision to localize a car on a network of roads by creating a road curvature descriptor. Existing literature also explored localization with respect to both visual and behavior landmarks~\cite{carloc}. Amazon Mechanical Turk (AMT) has been previously used for annotating vision data to obtain ground truth for training~\cite{Annotation2008} and for creating an image ontology database~\cite{ImageNet2009}. CrowdSearch~\cite{Yan:2010:CEC:1814433.1814443} is an image search system designed for mobile phones that employs AMT for crowdsourcing and human validation. It combines automated image search with real-time human validation of search results. In CamLoc, we use a similar system~\cite{Satyam} for gathering the human annotations.

% \textbf{Object Pose estimation} Iterative Closest Point (ICP)~\cite{ICT91,ICT92} and RANSAC ~\cite{Fischler:1981:RSC:358669.358692} are fast heuristics for object pose estimation (called registration). But, in our work, we are interested in camera pose estimation.

% \textbf{Image Sensor System} In Distributed Image Search~\cite{Yan:2008:DIS:1460412.1460428}, a user can query a distributed camera sensor system by specifying an image. This query goes to all sensors. The system returns the top $k$ most relevant matching sensors/images.

% \textbf{Sport Analytics} Sports analytics is a growing area of interest.
% \citet{Halvorsen:2013:BIS:2483977.2483982} present Bagadus, a video processing system that utilizes a video camera array for following a particular soccer player.
% \citet{tennis2013} describe a tennis player tracking system.
% \citet{Stein2018Sport} propose a visual analytics system that combines video and movement data for soccer player analysis.
% \marcos{This paragraph is in case we include sport analytics.}

%%% Local Variables:
%%% mode: latex
%%% TeX-master: "camloc"
%%% End:
%% \input{misc.tex}
\section{Conclusion}
\label{sec:conclusion}

This paper explores the following question: under what conditions is
it possible to estimate the location of a camera from a single image
taken by the camera? Through a set of carefully designed system
components we show that it is possible to use a combination of
crowd-sourced annotation from human workers, concepts of vanishing
point in projection geometry, the standard dimensions of an object in
the image, and the pixel to GPS mapping of two pixels, to uniquely
determine the position and height of the camera with less than 12
meters error in 95\% of the cases. We also discuss how the pixel to
GPS mapping of two pixels can be obtained for two particular cases
on real world scenarios. Through an instantiation of our ideas in a
tool called CamLoc, we show that our system performs two orders of
magnitude better than a state-of-the-art neural network based camera
localization system. We further show that Camloc can be used to implement different virtual 
sensors: virtual scale, virtual clinometer, virtual radar, and virtual guide.
Future work with CamLoc involves exploring more
complex real-world scenarios, developing techniques to obtain the
pixel to GPS mapping for such scenarios, and using multiple images to 
improve the error performance even further.

%% Finally, we plan to combine the relative
%% position estimation from images with other sensing modalities such as
%% radio signal strength information to associate the human/cars with the
%% the mobile device.

%%% Local Variables:
%%% mode: latex
%%% TeX-master: "camloc"
%%% End:

%%\section*{Acknowledgment}
%\newpage
\bibliographystyle{unsrtnat}
\bibliography{references.bib}
\end{document}